\documentclass[journal]{IEEEtran}
\usepackage[pdftex]{graphicx}
\usepackage{arydshln} % for dashed line
\usepackage{multirow}% http://ctan.org/pkg/hhline
\usepackage{hhline}
\usepackage{ehhline}
\usepackage{dashrule}
\usepackage{amssymb} 
\usepackage{pifont} 
\newcommand{\xmark}{\ding{55}}%

\usepackage{cite}
\usepackage{xcolor}

\ifCLASSINFOpdf
\else
\fi
\usepackage{amsmath}
\usepackage{textcomp}
\usepackage{color,soul} % highlighting etc

\usepackage{fixltx2e}
\usepackage{url}
\renewcommand{\thesubsection}{\Alph{subsection}}

\newcounter{sss}[subsubsection]
\newenvironment{sss}[1][]{\refstepcounter{sss}\thesss)}{}

\newcommand\cev[1]{\overleftarrow{#1}}
\renewcommand\vec[1]{\overrightarrow{#1}}

\begin{document}
\title{A Survey on Explainable Artificial Intelligence (XAI): towards Medical XAI}

\author{Erico~Tjoa,
        and~Cuntai~Guan,~\IEEEmembership{Fellow,~IEEE}
% <-this % stops a space
\thanks{Erico T. and Cuntai Guan were with the School of Computer Science and Engineering,  Nanyang Technological University, Singapore.}% <-this % stops a space
\thanks{Erico T. was also affiliated with  HealthTech Division, Alibaba Group Holding Limited.}}% <-this % stops a space}


\markboth{Journal of \LaTeX\ Class Files,~Vol.~14, No.~8, August~2015}%
{Shell \MakeLowercase{\textit{et al.}}: Bare Demo of IEEEtran.cls for IEEE Journals}



\maketitle

\begin{abstract}
Recently, artificial intelligence and machine learning in general have demonstrated remarkable performances in many tasks, from image processing to natural language processing, especially with the advent of deep learning. Along with research progress, they have encroached upon many different fields and disciplines. Some of them require high level of accountability and thus transparency, for example the medical sector. Explanations for machine decisions and predictions are thus needed to justify their reliability. This requires greater interpretability, which often means we need to understand the mechanism underlying the algorithms. Unfortunately, the blackbox nature of the deep learning is still unresolved, and many machine decisions are still poorly understood. We provide a review on interpretabilities suggested by different research works and categorize them. The different categories show different dimensions in interpretability research, from approaches that provide ``obviously" interpretable information to the studies of complex patterns. By applying the same  categorization to interpretability in medical research, it is hoped that (1) clinicians and practitioners can subsequently approach these methods with caution, (2) insights into interpretability will be born with more considerations for medical practices, and (3) initiatives to push forward data-based, mathematically- and technically-grounded medical education are encouraged.
\end{abstract}


% \fi
\IEEEpeerreviewmaketitle

\section{Introduction}
\IEEEPARstart{M}{achine learning} (ML) has grown large in both research and industrial applications, especially with the success of deep learning (DL) and neural networks (NN), so large that its impact and possible after-effects can no longer be taken for granted. In some fields, failure is not an option: even a momentarily dysfunctional computer vision algorithm in autonomous vehicle easily leads to fatality. In the medical field, clearly human lives are on the line. Detection of a disease at its early phase is often critical to the recovery of patients or to prevent the disease from advancing to more severe stages. While machine learning methods, artificial neural networks, brain-machine interfaces and related subfields have recently demonstrated promising performance in performing medical tasks, they are hardly perfect \cite{DiBrain, 
DBLP:journals/corr/RonnebergerFB15, 
10.1016/S1071-5819(03)00038-7,
CHEN2017633,
DBLP:journals/corr/CicekALBR16,
DBLP:journals/corr/abs-1901-07031,
DBLP:journals/corr/MilletariNA16,
journals/corr/ChenPK0Y16,
Kelly2019}.

Interpretability and explainability of ML algorithms have thus become pressing issues: who is accountable if things go wrong? Can we explain why things go wrong? If things are working well, do we know why and how to leverage them further? Many papers have suggested different measures and frameworks to capture interpretability, and the topic explainable artificial intelligence (XAI) has become a hotspot in ML research community. Popular deep learning libraries have started to include their own explainable AI libraries, such as Pytorch Captum and Tensorflow tf-explain.  Furthermore, the proliferation of interpretability assessment criteria (such as \textit{reliability}, \textit{causality} and \textit{usability}) helps ML community keep track of how algorithms are used and how their usage can be improved, providing guiding posts for further developments \cite{doshivelez2017towards,DBLP:journals/corr/abs-1905-05134,10.1145/358916.358995}. In particular, it has been demonstrated that visualization is capable of helping researchers detect erroneous reasoning in classification problems that many previous researchers possibly have missed \cite{Lapuschkin2019}. 

The above said, there seems to be a lack of uniform adoption of interpretability assessment criteria across the research community. There have been attempts to define the notions of ``interpretability", ``explainability" along with ``reliability", ``trustworthiness" and other similar notions without clear expositions on how they should be incorporated into the great diversity of implementations of machine learning models; consider \cite{doshivelez2017towards,10.1145/2939672.2939778,DBLP:journals/corr/Lipton16a, 8400040, 8631448, BARREDOARRIETA202082}. In this survey, we will instead use ``explainability" and ``interpretability" interchangeably, considering a research to be related to interpretability if it does show any attempts (1) to explain the decisions made by algorithms, (2) to uncover the patterns within the inner mechanism of an algorithm, (3) to present the system with coherent models or mathematics, and we will include even loose attempts to raise the credibility of machine algorithms.

In this work, we survey through research works related to the interpretability of ML or computer algorithms in general, categorize them, and then apply the same categories to interpretability in the medical field. The categorization is especially aimed to give clinicians and practitioners a perspective on the use of interpretable algorithms that are available in diverse forms. The trade-off between the ease of interpretation and the need for specialized mathematical knowledge may create a bias in preference for one method compared to another without justification based on medical practices. This may further provide a ground for specialized education in the medical sector that is aimed to realize the potentials that reside within these algorithms. We also find that many journal papers in the machine learning and AI community are algorithm-centric. They often assume that the algorithms used are obviously interpretable without conducting human subject tests to verify their interpretability; see column HSI of table \ref{tab:tablepart1} and \ref{tab:tablepart2}. Note that assuming that a model is obviously interpretable is not necessarily wrong, and, in some cases human tests might be irrelevant (for example pre-defined models based on commonly accepted knowledge specific to the content-subject may be considered interpretable without human subject tests). In the tables, we also include a column to indicate whether the interpretability method applies for artificial NN, since the issue of interpretability is recently gathering attention due to its blackbox nature.

We will not attempt to cover all related works many of which are already presented in the research papers and survey we cite \cite{DiBrain,DBLP:journals/corr/RonnebergerFB15,SOEKADAR2015172, doi:10.1002/widm.1312 , DBLP:journals/corr/abs-1902-06019 , DBLP:journals/corr/Lipton16a , Vellido2019 , Topol2019 , 8610271 , Kallianos2019 , MONTAVON20181 , DBLP:journals/corr/abs-1708-08296 , Rieger2018, 10.1007/978-3-030-22871-2_67, 8400040, 8631448, 8889997,BARREDOARRIETA202082}. We extend the so-called \textit{integrated interpretability} \cite{8400040} by including considerations for subject-content-dependent models. Compared to \cite{8631448}, we also overview the mathematical formulation of common or popular methods, revealing the great variety of approaches to interpretability. Our categorization draws a starker borderline between the different views of interpretability that seem to be difficult to reconcile. In a sense, our survey is more suitable for technically-oriented readers due to some mathematical details, although casual readers may find useful references for relevant popular items, from which they may develop interests in this young research field. Conversely, algorithm users that need interpretability in their work might develop an inclination to understand what is previously hidden in the thick veil of mathematical formulation, which might ironically undermine reliability and interpretability. Clinicians and medical practitioners already having some familiarity with mathematical terms may get a glimpse on how some proposed interpretability methods might be risky and unreliable. The survey \cite{8889997} views interpretability in terms of extraction of relational knowledge, more specifically, by scrutinizing the methods under \textit{neural-symbolic cycle}. It presents the framework as a sub-category within the interpretability literature. We include it under \textit{verbal interpretability}, though the framework does demonstrate that methods in other categories can be perceived under verbal interpretability as well. The extensive survey \cite{BARREDOARRIETA202082} provides a large list of researches categorized under \textit{transparent model} and models requiring \textit{post-hoc analysis} with multiple sub-categories. Our survey, on the other hand, aims to overview the state of interpretable machine learning as applied to the medical field.

This paper is arranged as the following. Section II introduces generic types of interpretability and their sub-types. In each section, where applicable, we provide challenges and future prospects related to the category. Section III applies the categorization of interpretabilities in section II to medical field and lists a few risks of machine interpretability in the medical field. Before we proceed, it is also imperative to point out that the issue of accountability and interpretability has spawned discussions and recommendations \cite{canweopenblackbox,doi:10.1002/hbm.24886,brundage2020trustworthy}, and even entered the sphere of ethics and law enforcements \cite{eu_xai_guideline}, engendering movements to protect the society from possible misuses and harms in the wake of the increasing use of AI.

\begin{figure}[h!]
\centering
\includegraphics[width=3.4in, trim = {0 0 0 0.5cm}]{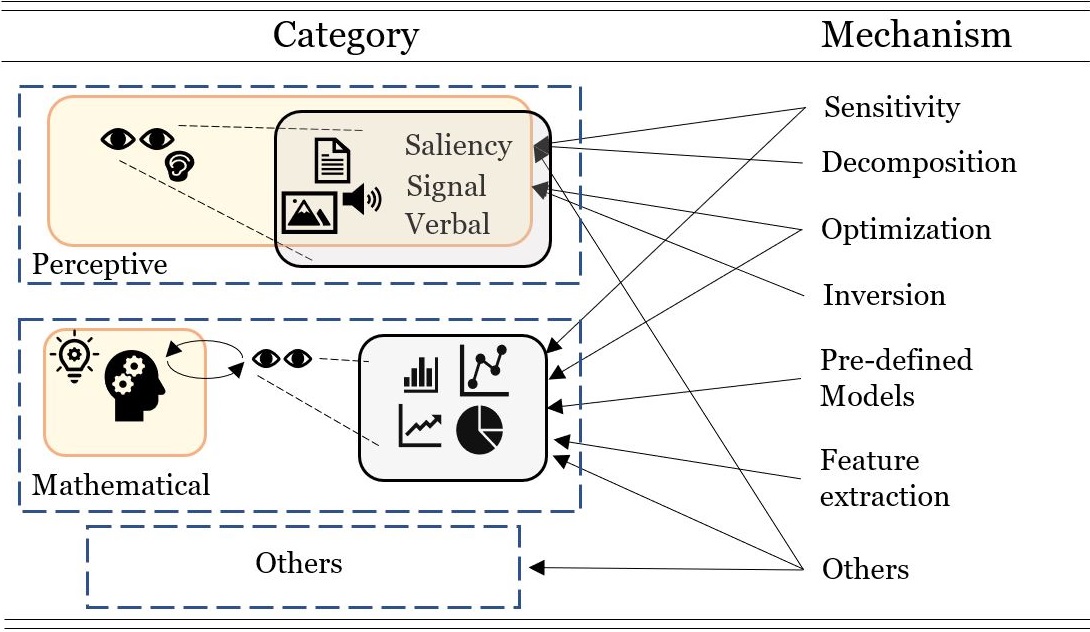}
\caption{Overview of categorization with illustration. Orange box: interpretability interface to demarcate the separation between interpretable information and the cognitive process required to understand them. Grey box: algorithm output/product that is proposed to provide interpretability. Black arrow: computing or comprehension process. The perceptive interpretability methods generate items that are usually considered immediately interpretable. On the other hand, methods that provide interpretability via mathematical structure generate outputs that require one more layer of cognitive processing interface before reaching the interpretable interface. The eyes and ear icons represent human senses interacting with items generated for interpretability. }
\label{OverviewFigure}
\end{figure}

\section{Types of Interpretability}
There has yet to be a widely-adopted standard to understand ML interpretability, though there have been works proposing frameworks for interpretability \cite{doshivelez2017towards,Lapuschkin2019,10.1145/3290605.3300831}. In fact, different works use different criteria, and they are justifiable in one way or another. Reference \cite{8099837} suggests \textit{network dissection} for the interpretability of visual representations and offers a way to quantify it as well. The interactive websites \cite{Olah_2017,olah_buildingblock} have suggested a unified framework to study interpretabilities that have thus-far been studied separately. The paper \cite{NIPS2017_7062} defines a unified measure of \textit{feature importance} in the SHAP (SHapley Additive exPlanations) framework. Here, we categorize existing interpretabilities and present a non-exhaustive list of works in each category.

The two major categories presented here, namely \textit{perceptive interpretability} and \textit{interpretability by mathematical structures}, as illustrated in fig. {\ref{OverviewFigure}}, appear to present different polarities within the notion of interpretability. An example of the difficulty with perceptive interpretability is as the following. When a visual ``evidence" is given erroneously, the algorithm or method used to generate the ``evidence" and the underlying mathematical structure sometimes do not offer any useful clues on how to fix the mistakes. On the other hand, a mathematical analysis of patterns may provide information in high dimensions. They can only be easily perceived once the pattern is brought into lower dimensions, abstracting some fine-grained information we could not yet prove is not discriminative with measurable certainty.

\begin{figure*}[h]
\centering
\includegraphics[width=5.8in]{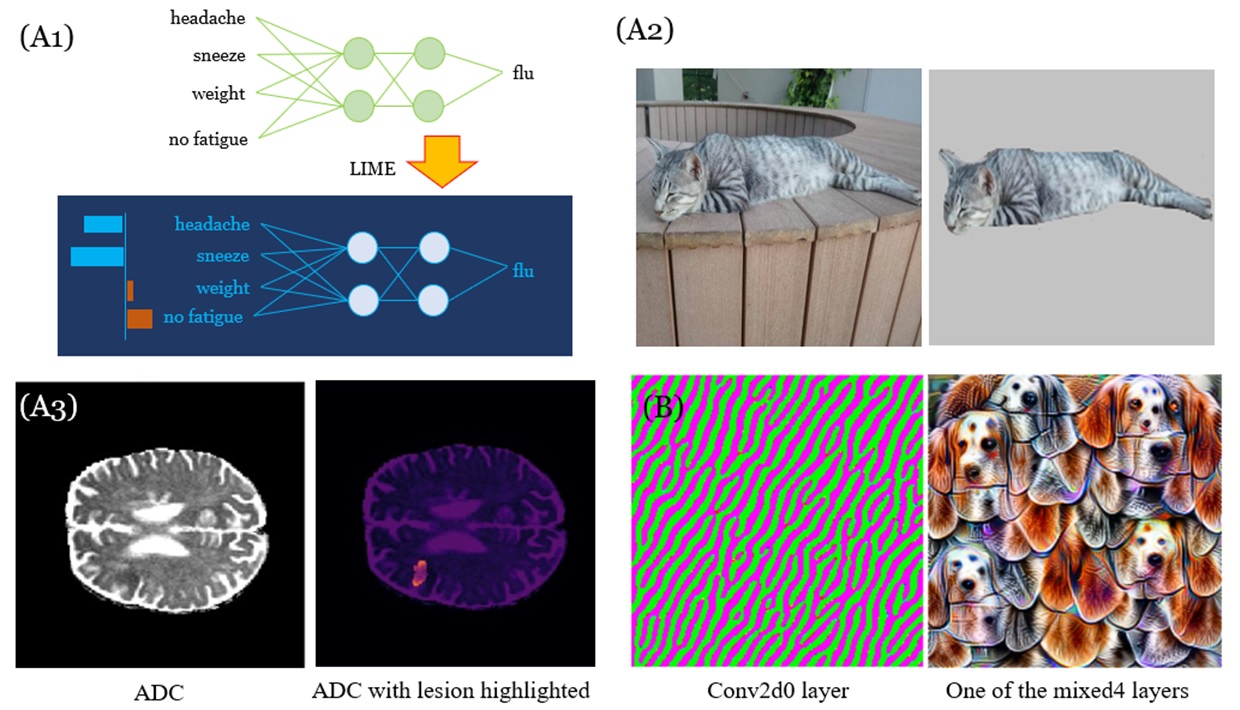}
\caption{(A1) Using LIME to generate explanation for text classification. \textit{Headache} and \textit{sneeze} are assigned positive values. This means both factors have positive contribution to the model prediction “flu”. On the other hand, \textit{weight} and \textit{no fatigue} contribute negatively to the prediction. (A2) LIME is used to generate the super-pixels for the classification “cat”. (A3) ADC modality of a slice of MRI scan from ISLES 2017 segmentation competition. Reddish intensity region reflects a possible “explanation” to the choice of segmentation (segmentation not shown) (B) Optimized images that maximize the activation of a neuron in the indicated layers. In shallower layer, simple patterns activate neurons strongly while in deeper layer, more complex features such as dog faces and ears do. Figure (B) is obtained from \protect\url{https://distill.pub/2018/building-blocks/} with permission from Chris Olah.}
\label{fig:typesXAI}
\end{figure*}

\subsection{Perceptive Interpretability} \label{percept}
\begin{figure*}[h]
\centering
\includegraphics[width=7.2in]{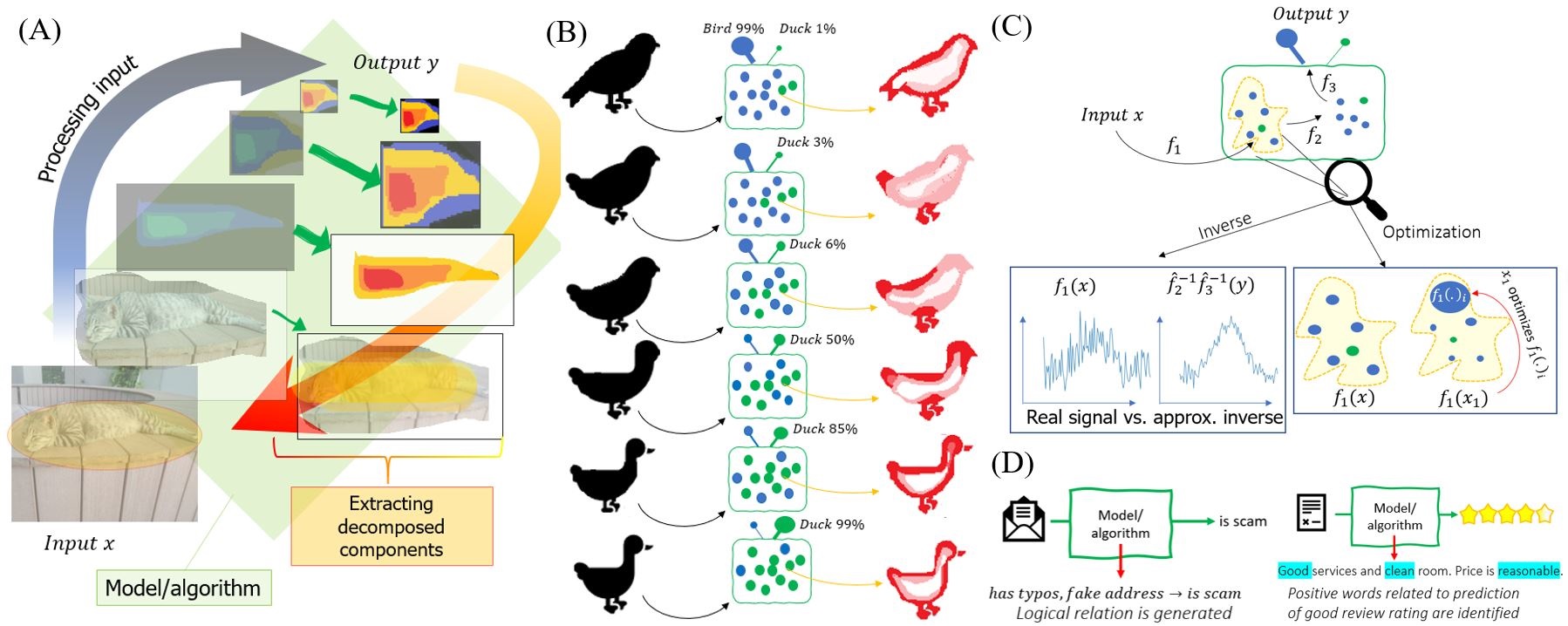}
\caption{Overview on perceptive interpretability methods. (A) \textit{Saliency method with decomposition mechanism}. The input which is an image of a cat is fed into the model for processing along the blue arrow. The resulting output and intermediate signals (green arrows) are decomposed and selectively picked for processing, hence providing information for the intermediate mechanism of the model in the form of (often) heatmappings, shown in red/orange/yellow colors. (B) \textit{Saliency method with sensitivity mechanism}. The idea is to show how small changes to the input (black figures of birds and ducks) affect the information extracted for explainability (red silhouette). In this example, red regions indicate high relevance, which we sometimes observe at edges or boundary of objects, where gradients are high. (C) \textit{Signal method by inversion and optimization}. Inverses of signals or data propagated in a model could possibly reveal more sensible information (see arrow labeled ``Inverse"). Adjusting input to optimize a particular signal (shown as the i-th component of the function \(f_1\)) may provide us with \(x_1\) that reveals explainable information (see arrow labeled ``Optimizatin"). For illustration, we show that the probability of correctly predicting duck improves greatly once the head is changed to the head of a duck which the model recognizes. (D) \textit{Verbal interpretability} is typically achieved by ensuring that the model is capable of providing humanly understable statements, such as the logical relation or the positive words shown.}
\label{fig:overviewPerc}
\end{figure*}

We include in this category interpretabilities that can be humanly perceived, often one that will be considered “obvious”. For example, as shown in fig. \ref{fig:typesXAI}(A2), an algorithm that classifies an image into the “cat” category can be considered obviously interpretable if it provides segmented patch showing the cat as the explanation. We should note that this alone might on the other hand be considered insufficient, because (1) it still does not un-blackbox an algorithm and (2) it ignores the possibility of using background objects for its decision. The following are the sub-categories to perceptive interpretability. Refer to fig. \ref{fig:overviewPerc} for the overview of the common sub-categories.

\textit{\thesubsection .\begin{sss}\end{sss} Saliency}

Saliency method explains the decision of an algorithm by assigning values that reflect the importance of input components in their contribution to that decision. These values could take the forms of probabilities and super-pixels such as heatmaps etc. For example, fig. \ref{fig:typesXAI}(A1) shows how a model predicts that the patient suffers from flu from a series of factors, but LIME \cite{10.1145/2939672.2939778} explains the choice by highlighting the importance of the particular symptoms that indicate that the illness should indeed be flu. Similarly, \cite{DBLP:journals/corr/abs-1809-08037} computes the scores reflecting the n-grams activating convolution filters in NLP (Natural Language Processing). Fig. \ref{fig:typesXAI}(A2) demonstrates the output that LIME will provide as the explanation for the choice of classifications “cat” and fig. \ref{fig:typesXAI}(A3) demonstrates a kind of heatmap that shows the contribution of pixels to the segmentation result (segmentation result not shown, and this figure is only for demonstration). More formally, given that model \(f\) makes a prediction \(y=f(x)\) for input \(x\), for some metric \(v\), typically large magnitude of \(v(x_{i})\) indicates that the component \(x_{i}\) is a significant reason for the output \(y\). 

Saliency methods via decomposition have been developed. In general, they decompose signals propagated within their algorithm and selectively rearrange and process them to provide interpretable information. Class Activation Map (CAM) has been a popular method to generate heat/saliency/relevance-map (from now, we will use the terms interchangeably) that corresponds to discriminative features for classifications \cite{CAM7780688, DBLP:journals/corr/SelvarajuDVCPB16, 10.1007/978-3-030-00928-1_55}. The original implementation of CAM \cite{CAM7780688} produces heatmaps using \(f_k (x,y)\), the pixel-wise activation of unit \(k\) across spatial coordinates \((x,y)\) in the last convolutional layers, weighted by \(w^c_k\), the coefficient corresponding to unit \(k\) for class \(c\). CAM at pixel \((x,y)\) is thus given by \(M_c(x,y)=\Sigma_k w_k^c f_k(x,y)\). 

Similarly, widely used Layer-wise Relevance Propagation (LRP) is introduced in \cite{10.1371/journal.pone.0130140}. Some papers that use LRP to construct saliency maps for interpretability include \cite{evalLRP2017,Lapuschkin2019,DBLP:journals/corr/abs-1807-03418,DBLP:journals/corr/abs-1810-09945,DBLP:journals/corr/ArrasHMMS16a,InterpretableHumanAction,eberle2020building}. It is also applicable for video processing \cite{hiley2020explaining}. A short summary for LRP is given in \cite{samek2016interpreting}. LRP is considered a decomposition method \cite{heatmappingWCCI}. Indeed, the importance scores are decomposed such that the sum of the scores in each layer will be equal to the output. In short, the relevance score is the pixel-wise intensity at the input layer \(R^{(0)}\) where \(R_i^{(l)}=\Sigma_j \frac{a_i^{(l)} w_{ij}^+}{\Sigma_i a_i^{(l)} w_{ij}^+} R_j^{(l+1)}\) is the relevance score of neuron \(i\) at layer \(l\) with the input layer being at \(l=0\). Each pixel \((x,y)\) at the input layer is assigned the importance value \(R^{(0)}(x,y)\), although some combinations of relevance scores \(\{R^{(l)}_c\}\) at inner layer \(l\) over different channels \(\{c\}\) have been demonstrated to be meaningful as well (though possibly less precise; see the tutorial in its website heatmapping.org). LRP can be understood in Deep Taylor Decomposition framework \cite{DeepTaylorDecompNN}. The code implementation can also be found in the aforementioned website.

Automatic Concept-based Explanations (ACE) algorithm \cite{ACE_NIPS2019_9126} uses super-pixels as explanations. Other decomposition methods that have been developed include, DeepLIFT and gradient*input \cite{DBLP:journals/corr/ShrikumarGK17}, Prediction Difference Analysis \cite{DBLP:journals/corr/ZintgrafCAW17} and \cite{DBLP:journals/corr/abs-1809-08037}. Peak Response Mapping \cite{CVPR_PRM} is generated by backpropagating peak signals. Peak signals are normalized and treated as probability, and the method can be seen as decomposition into probability transitions. In \cite{kindermans2017learning}, \textit{Removed correlation} \(\rho\) is proposed as a metric to measure the quality of signal estimators. And then it proposes PatternNet and PatternAttribution that backpropagate parameters optimized against \(\rho\), resulting in saliency maps as well. SmoothGrad \cite{smilkov2017smoothgrad} improves gradient-based techniques by adding noises. Do visit the related website that displays numerous visual comparison of saliency methods; be mindful of how some heatmaps highlight apparently irrelevant regions. 

For natural language processing or sentiment analysis, saliency map can also take the form of “heat” scores over words in texts, as demonstrated by \cite{arras-etal-2017-explaining} using LRP and by \cite{karpathy2015visualizing}. In the medical field (see later section), \cite{DBLP:journals/corr/abs-1901-07031, 10.1007/978-3-030-00928-1_55, 10.1007/978-3-030-00928-1_56, 10.1007/978-3-030-00934-2_29, 10.1007/978-3-030-00931-1_24,DBLP:journals/corr/abs-1805-08403,Tang2019NatureAlz,10.1117/12.2549298, 10.1007/978-3-030-33850-3_3} have studied methods employing saliency and visual explanations. Note that we also sub-categorize LIME as a method that uses optimization and sensitivity as its underlying mechanisms, and many researches on interpretability span more than one sub-categories. 

\textit{Challenges and Future Prospects}. As seen, the formulas for CAM and LRP are given on a heuristic: certain ways of interaction between weights and the strength of activation of some units within the models will eventually produce the interpretable information. The intermediate processes are not amenable to scrutiny. For example, taking one of the weights and changing its value does not easily reveal any useful information. How these prescribed ways translate into interpretable information may also benefit from stronger evidences, especially evidences beyond visual verification of localized objects. Signal methods to investigate ML models (see later section) exist, but such methods that probe them with respect to the above methods have not been attempted systematically, possibly opening up a different research direction.

\textit{\thesubsection .\begin{sss}\end{sss} Signal Method}

Methods of interpretability that observe the stimulation of neurons or a collection of neurons are called signal methods \cite{Kindermans2019Unreliability}. On the one hand, the activated values of neurons can be manipulated or transformed into interpretable forms. For example, the activation of neurons in a layer can be used to reconstruct an image similar to the input. This is possible because neurons store information systematically \cite{DBLP:journals/corr/ZeilerF13}: \textit{feature maps} in the deeper layer activate more strongly to complex features, such as human face, keyboard etc while feature maps in the shallower layers show simple patterns such as lines and curves. Note: an example of \textit{feature map} is the output of a convolutional filter in a Convolutional Neural Network (CNN). On the other hand, parameters or even the input data might be optimized with respect to the activation values of particular neurons using methods known as \textit{activation optimization} (see a later section). The following are the relevant sub-categories.

\textit{Feature maps and Inversions for Input Reconstructions}. A feature map often looks like a highly blurred image with most region showing zero (or low intensity), except for the patch that a human could roughly discern as a detected feature. Sometimes, these discernible features are considered interpretable, as in \cite{DBLP:journals/corr/ZeilerF13}. However, they might be too distorted. 

Then, how else can a feature map be related to a humanly-perceptible feature? An inverse convolution map can be defined: for example, if feature map in layer 2 is computed in the network via \(y_2=f_2(f_1(x))\) where \(x\) is the input, \(f_1(.)\) consists of 7x7 convolutions of stride 2 followed by max-pooling and likewise \(f_2(.)\). Then \cite{DBLP:journals/corr/ZeilerF13} reconstructs an image using a deconvolution network by approximately inversing the trained convolutional network \(\tilde{x}=deconv(y)=\hat{f}_2^{-1}\hat{f}_1^{-1}(y)\) which is an approximation, because layers such as max-pooling have no unique inverse. It is shown that \(\tilde{x}\) does appear like slightly blurred version of the original image, which is distinct to human eye. Inversion of image representations within the layers has also been used to demonstrate that CNN layers do store important information of an input image accurately \cite{DBLP:journals/corr/MahendranV14,DBLP:journals/corr/DosovitskiyB15}. Guided backpropagation \cite{springenberg2014striving} modifies the way backpropagation is performed to achieve inversion by zeroing negative signals from both the output or input signals backwards through a layer. Indeed, inversion-based methods do use saliency maps for visualization of the “activated” signals.

\textit{Activation Optimization}. Besides transforming the activation of neurons, signal method also includes finding input images that optimize the activation of a neuron or a collection of neurons. This is called the \textit{activation maximization}. Starting with a noise as an input \(x\), the noise is slowly adjusted to increase the activation of a select (collection of) neuron(s) \(\{a_k\}\). In simple mathematical terms, the task is to find \(x_0=argmax\ ||\{a_k\}||\) where optimization is performed over input \(x\) and \(||.||\) is a suitable metric to measure the combined strength of activations. Finally the optimized input that maximizes the activation of the neuron(s) can emerge as something visually recognizable. For example, the image could be a surreal fuzzy combination of swirling patterns and parts of dog faces, as shown in fig. \ref{fig:typesXAI}(B). 

Research works on activation maximization include \cite{visualization_techreport} on MNIST dataset, \cite{DBLP:journals/corr/NguyenYC16} and \cite{DBLP:journals/corr/YosinskiCNFL15} that uses a regularization function. In particular, \cite{Olah_2017} provides an excellent interactive interface (feature visualization) demonstrating activation-maximized images for GoogLeNet \cite{googleNet}. GoogLeNet has a deep architecture, from which we can see how neurons in deeper layer stores complex features while shallower layer stores simple patterns; see fig. \ref{fig:typesXAI}(B). To bring this one step further, the “semantic dictionary” is used \cite{olah_buildingblock} to provide a visualization of activations within a higher-level organization and semantically more meaningful arrangements. 

\textit{Other Observations of Signal Activations.} Ablation studies \cite{DBLP:journals/corr/abs-1901-08644,meyes2020hood} also study the roles of neurons in shallower and deeper layers. In essence, some neurons are corrupted and the output of the corrupted neural network is compared to the original network.

\textit{Challenges and Future Prospects}. Signal methods might have revealed some parts of the black-box mechanisms. Many questions still remain. 
\begin{itemize}
\item What do we do with the (partially) reconstructed images and images that optimize activation? 
\item We might have learned how to approximately inverse signals to recover images, can this help improve interpretability further? 
\item The components and parts in the intermediate process that reconstruct the approximate images might contain important information; will we be able to utilize them in the future? 
\item How is explaining the components in this ``inverse space'' more useful than explaining signals that are forward propagated? 
\item Similarly, how does looking at intermediate signals that lead to activation optimization help us pinpoint the role of a collection of neurons? 
\item Optimization of highly parameterized functions notoriously gives non-unique solutions. Can we be sure that optimization that yields combination of surreal dog faces will not yield other strange images with minor alteration? 
\end{itemize}

In the process of answering these questions, we may find hidden clues required to get closer to interpretable AI. 

\textit{\thesubsection .\begin{sss}\end{sss} Verbal Interpretability}

This form of interpretability takes the form of verbal chunks that human can grasp naturally. Examples include sentences that indicate causality, as shown in the examples below.

Logical statements can be formed from proper concatenation of predicates, connectives etc.  An example of logical statement is the conditional statement. Conditional statements are statements of the form \(A\rightarrow B\), in another words \textit{“if A then B”}. An ML model from which logical statements can be extracted directly has been considered obviously interpretable. The survey \cite{8889997} shows how interpretability methods in general can be viewed under such symbolic and relational system. In the medical field, see for example \cite{10.1145/2783258.2788613,Letham_2015}.

Similarly, \textit{decision sets} or \textit{rule sets} have been studied for interpretability  \cite{DBLP:journals/corr/abs-1902-00006}. The following is a single line in a rule set ``rainy and grumpy or calm \(\rightarrow\) dairy or vegetables'', directly quoted from the paper. Each line in a rule set contains a clause with an input in \textit{disjunctive normal form} (DNF) mapped to an output in DNF as well. The example above is formally written (rainy\(\wedge\)grumpy)\(\vee\)calm\(\rightarrow\)dairy\(\vee\)vegetables. Comparing three different variables, it is suggested that interpretability of explanations in the form of rule sets is most affected by cognitive chunks, explanation size and little effected by variable repetition. Here, a cognitive chunk is defined as a clause of inputs in DNF and the number of (repeated) cognitive chunks in a rule set is varied. The explanation size is self-explanatory (a longer/shorter line in a rule set, or more/less lines in a rule set). MUSE \cite{museHima} also produces explanation in the form of decision sets, where interpretable model is chosen to approximate the black-box function and optimized against a number of metrics, including direct optimization of interpretability metrics.

It is not surprising that verbal segments are provided as the explanation in NLP problems. An encoder-generator framework \cite{lei-etal-2016-rationalizing} extracts segment like ``a very pleasant ruby red-amber color'' to justify 5 out of 5-star rating for a product review. Given a sequence of words \(x=(x_1,...,x_l)\) with \(x_k\in\mathbb{R}^d\), explanation is given as the subset of the sentence  that gives a summary of why the rating is justified. The subset can be expressed as the binary sequence \((z_1,...,z_l)\) where \(z_k=1 (0)\) indicates \(x_k\) is (not) in the subset. Then \(z\) follows a probability distribution with \(p(z|x)\) decomposed by assuming independence to \(\Pi_k p(z_k |x)\) where \(p(z_k|x)=\sigma_z (W^z[\vec{h_k},\cev{h_k}]+b^z)\), with \(\vec{h_t},\cev{h_t}\) being the usual hidden units in the recurrent cell (forward and backward respectively). Similar segments are generated using filter-attribute probability density function to improve the relation between the activation of certain filters and specific attributes \cite{DBLP:journals/corr/abs-1805-08969}. Earlier works on Visual Question Answering (VQA) \cite{10.1007/s11263-016-0966-6,10.5555/3157096.3157129,Das_2017} are concerned with the generation of texts discussing objects appearing in images. 

\textit{Challenges and Future Prospects}. While texts appear to provide explanations, the underlying mechanisms used to generate the texts are not necessarily explained. For example, NNs and the common variants/components used in text-related tasks such as RNN (recurrent NN), LSTM (long short term memory) are still black-boxes that are hard to troubleshoot in the case of wrong predictions. There have been less works that probe into the inner signals of LSTM and RNN neural networks. This is a possible research direction, although similar problem as mentioned in the previous sub-subsection may arise (what to do with the intermediate signals?). Furthermore, while word embedding is often optimized with the usual loss minimization, there does not seem to be a coherent explanation to the process and shape of the optimized embedding. There may be some clues regarding optimization residing within the embedding, and thus successfully interpreting the shape of embedding may help shed light into the mechanism of the algorithm.

\begin{table*}[ht]
\caption{List of journal papers arranged according to the interpretability methods used, how interpretability is presented or the suggested means of interpretability. The tabulation provides a non-exhaustive overview of interpretability methods, placing some derivative methods under the umbrella of the main methods they derive from. HSI: Human Study on Interpretability \checkmark means there is human study designed to verify if the suggested methods are interpretable by the human subject. ANN: \checkmark means explicitly introduces new artificial neural network architecture, modifies existing networks or performs tests on neural networks.}
\begin{center}
\begin{tabular}{lcc|c|c|c}
    \hline
	Methods & HSI & ANN & Mechansim & & \\
    \hline
   
    CAM with global average pooling \cite{CAM7780688,10.1007/978-3-030-00934-2_34} & \xmark & \checkmark & \multirow{15}{*}{Decomposition} & \multirow{22}{*}{\rotatebox[origin=c]{-90}{Saliency}} & \multirow{36}{*}{\rotatebox[origin=c]{-90}{Perceptive Interpretability}}\\
    + Grad-CAM \cite{DBLP:journals/corr/SelvarajuDVCPB16} generalizes CAM, utilizing gradient & \checkmark & \checkmark & & & \\
    + Guided Grad-CAM and Feature Occlusion \cite{Tang2019NatureAlz} & \xmark & \checkmark & & & \\
    + Respond CAM \cite{10.1007/978-3-030-00928-1_55} & \xmark & \checkmark & & & \\
    + Multi-layer CAM \cite{NeuralMachineTrans} & \xmark & \checkmark & & & \\
    
    LRP (Layer-wise Relevance Propagation) \cite{Lapuschkin2019,samek2016interpreting} & \xmark & \checkmark &  & & \\
    + Image classifications. PASCAL VOC 2009 etc \cite{10.1371/journal.pone.0130140} & \xmark & \checkmark & & & \\
    + Audio classification. AudioMNIST \cite{DBLP:journals/corr/abs-1807-03418} & \xmark & \checkmark & & & \\
    + LRP on DeepLight. fMRI data from Human Connectome Project \cite{DBLP:journals/corr/abs-1810-09945} & \xmark & \checkmark & & & \\
    + LRP on CNN and on BoW(bag of words)/SVM \cite{DBLP:journals/corr/ArrasHMMS16a} & \xmark & \checkmark & & & \\
    + LRP on compressed domain action recognition algorithm \cite{InterpretableHumanAction} & \xmark & \xmark & & & \\
    + LRP on video deep learning, \textit{selective relevance method} \cite{hiley2020explaining} & \xmark & \checkmark & & & \\
    + BiLRP \cite{eberle2020building} & \xmark & \checkmark & & & \\   
    
	 DeepLIFT \cite{DBLP:journals/corr/ShrikumarGK17} & \xmark & \checkmark & & & \\ 
	 Prediction Difference Analysis \cite{DBLP:journals/corr/ZintgrafCAW17} & \xmark & \checkmark & & & \\ 
	 Slot Activation Vectors \cite{DBLP:journals/corr/abs-1809-08037} & \xmark & \checkmark & & & \\ 
	 PRM (Peak Response Mapping) \cite{CVPR_PRM} & \xmark & \checkmark & & & \\ 
	 	
	\hhline{%
	!{\leaders\hbox{\hdashrule{0.1 cm}{0.2pt}{1pt}}\hfil}%
	!{\leaders\hbox{\hdashrule{0.1 cm}{0.2pt}{1pt}}\hfil}%
	!{\leaders\hbox{\hdashrule{0.1 cm}{0.2pt}{1pt}}\hfil}%
	!{\leaders\hbox{\hdashrule{0.1 cm}{0.2pt}{1pt}}\hfil}%
	!{}%
	}
	%\hhline{-|-|-|-|~|~|}
	
    LIME (Local Interpretable Model-agnostic Explanations) \cite{10.1145/2939672.2939778} & \checkmark & \checkmark & \multirow{4}{*}{Sensitivity} & 		 &  \\
    + MUSE with LIME \cite{museHima}  & \checkmark & \checkmark & & & \\
    + Guidelinebased Additive eXplanation optimizes complexity, similar to LIME \cite{10.1007/978-3-030-33850-3_5} & \checkmark & \checkmark & & & \\
   \# Also listed elsewhere: \cite{Kindermans2019Unreliability,10.5555/3305890.3306024,10.1117/12.2549298,ACE_NIPS2019_9126} & N.A. & N.A. & & & \\

	\hhline{%
	!{\leaders\hbox{\hdashrule{0.1 cm}{0.2pt}{1pt}}\hfil}%
	!{\leaders\hbox{\hdashrule{0.1 cm}{0.2pt}{1pt}}\hfil}%
	!{\leaders\hbox{\hdashrule{0.1 cm}{0.2pt}{1pt}}\hfil}%
	!{\leaders\hbox{\hdashrule{0.1 cm}{0.2pt}{1pt}}\hfil}%
	!{}%
	}	    
     Others. Also listed elsewhere: \cite{ghorbani2017interpretation} & N.A. & N.A. & \multirow{4}{*}{Others} & & \\
     + Direct output labels. Training NN via multiple instance learning \cite{10.1007/978-3-030-00934-2_29} & \xmark & \checkmark & & & \\ 
     + Image corruption and testing Region of Interest statistically \cite{10.1007/978-3-030-00931-1_24} & \xmark & \checkmark & & & \\ 
     + Attention map with autofocus convolutional layer \cite{DBLP:journals/corr/abs-1805-08403} & \xmark & \checkmark & & & \\

    \hhline{-|-|-|-|-|~|}
    DeconvNet \cite{DBLP:journals/corr/ZeilerF13} & \xmark & \checkmark & \multirow{4}{*}{Inversion} & \multirow{9}{*}{\rotatebox[origin=c]{-90}{Signal}} & \\
    Inverting representation with natural image prior \cite{DBLP:journals/corr/MahendranV14} & \xmark & \checkmark &  &  & \\
    Inversion using CNN \cite{DBLP:journals/corr/DosovitskiyB15} & \xmark & \checkmark &  &  & \\
    Guided backpropagation \cite{springenberg2014striving,10.1007/978-3-030-00934-2_34} & \xmark & \checkmark &  &  & \\

	\hhline{%
	!{\leaders\hbox{\hdashrule{0.1 cm}{0.2pt}{1pt}}\hfil}%
	!{\leaders\hbox{\hdashrule{0.1 cm}{0.2pt}{1pt}}\hfil}%
	!{\leaders\hbox{\hdashrule{0.1 cm}{0.2pt}{1pt}}\hfil}%
	!{\leaders\hbox{\hdashrule{0.1 cm}{0.2pt}{1pt}}\hfil}%
	!{}%
	}	
	Activation maximization/optimization \cite{Olah_2017} & \xmark & \checkmark & \multirow{5}{*}{Optimization}&  &  \\
 + Activation maximization on DBN (Deep Belief Network) \cite{visualization_techreport} & \xmark & \checkmark &  &  &  \\
 + Activation maximization, multifaceted feature visualization \cite{DBLP:journals/corr/NguyenYC16} & \xmark & \checkmark & &  &  \\
 Visualization via regularized optimization \cite{DBLP:journals/corr/YosinskiCNFL15} & \xmark & \checkmark & &  &  \\
 Semantic dictionary \cite{olah_buildingblock} & \xmark & \checkmark & &  & \\ 

\hhline{-|-|-|-|-|~|}

    Decision trees & N.A. & N.A. & \multicolumn{2}{c|}{} & \\   
	Propositional logic, rule-based \cite{10.1145/2783258.2788613} & \xmark & \xmark & \multicolumn{2}{c|}{} &  \\
	Sparse decision list \cite{Letham_2015} & \xmark & \xmark & \multicolumn{2}{c|}{}  &  \\
	Decision sets, rule sets \cite{DBLP:journals/corr/abs-1902-00006,museHima} & \checkmark & \xmark & \multicolumn{2}{c|}{Verbal}  &  \\
	Encoder-generator framework \cite{lei-etal-2016-rationalizing} & \xmark & \checkmark & \multicolumn{2}{c|}{}  &  \\
	Filter Attribute Probability Density Function \cite{DBLP:journals/corr/abs-1805-08969} & \xmark & \xmark & \multicolumn{2}{c|}{}  &  \\
	MUSE (Model Understanding through Subspace Explanations) \cite{museHima} & \checkmark & \checkmark & \multicolumn{2}{c|}{}  &  \\

    \hline
\end{tabular}
\end{center}
\label{tab:tablepart1}
\end{table*}

\subsection{Interpretability via Mathematical Structure}
\begin{figure}[h]
\centering
\includegraphics[width=3.6in]{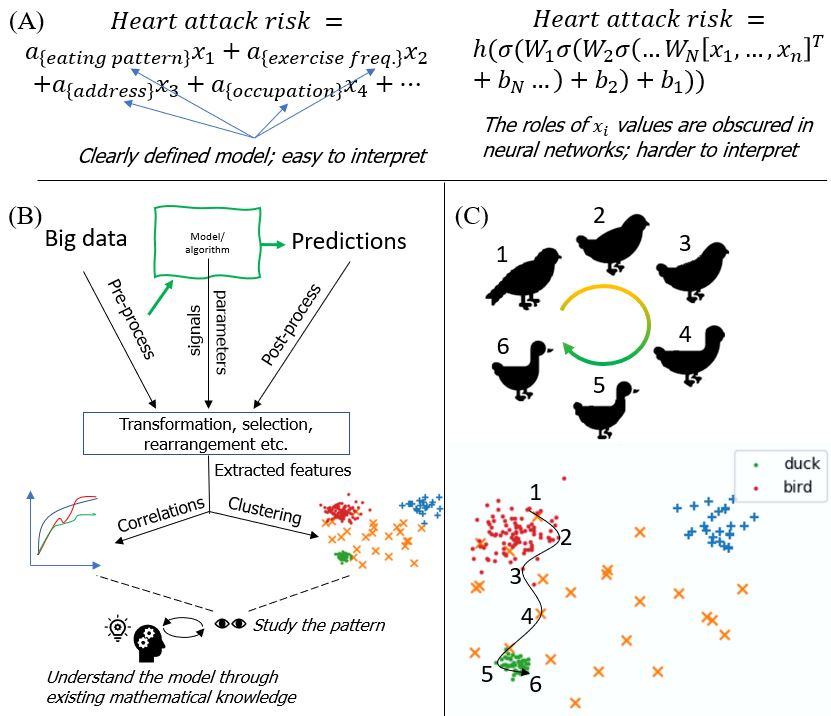}
\caption{Overview of methods whose interpretability depend on interpreting underlying mathematical structure. (A) \textit{Pre-defined models}. Modeling with clear, easily understandable model, such as linear model can help improve readability, and hence interpretability. On the other hand, using neural network could obscure the meaning of input variables. (B) \textit{Feature extraction}. Data, predicted values, signals and parameters from a model are processed, transformed and selectively picked to provide useful information. Mathematical knowledge is usually required to understand the resulting pattern. (C) \textit{Sensitivity}. Models that rely on sensitivity, gradients, perturbations and related concepts will try to account for how different data are differently represented. In the figure, the small changes transforming the bird to the duck can be traced along a map obtained using clustering.}
\label{fig:overviewMath}
\end{figure}

\begin{figure*}[h]
\centering
\includegraphics[width=6.4in]{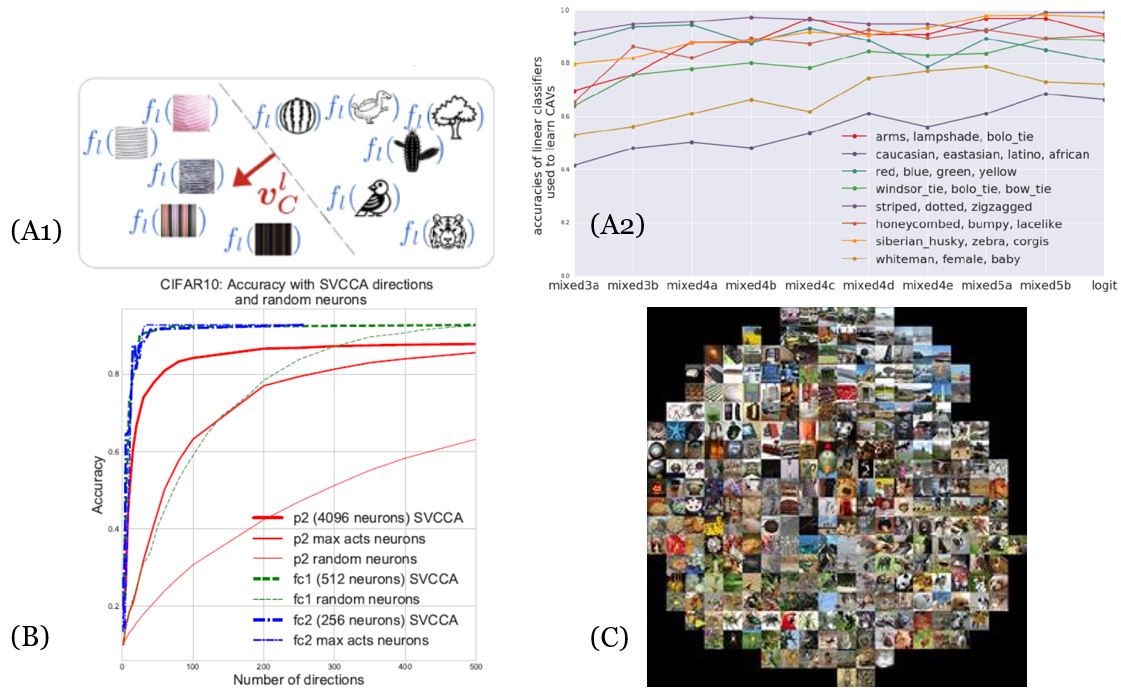}
\caption{(A1) TCAV \cite{conf_icml_KimWGCWVS18} method finds the hyperplane CAV that separates concepts of interest. (A2) Accuracies of CAV applied to different layers supports the idea that deeper NN layers contain more complex concepts, and shallower layers contain simpler concepts. (B) SVCCA \cite{SVCCARaghu} finds the most significant subspace (direction) that contains the most information. The graph shows that as few as 25 directions out of 500 are enough to produce the accuracies of the full network. (C) t-SNE clusters images in meaningful arrangement, for example dog images are close together. Figures (A1,A2) are used with permission from the authors Been Kim; figures (B,C) from Maithra Raghu and Jascha Sohl-dickstein.}
\label{fig:XAIMath}
\end{figure*}

Mathematical structures have been used to reveal the mechanisms of ML and NN algorithms. In the previous section, deeper layer of NN is shown to store complex information while shallower layer stores simpler information \cite{DBLP:journals/corr/ZeilerF13}. TCAV \cite{conf_icml_KimWGCWVS18} has been used to show similar trend, as suggested by fig. \ref{fig:XAIMath}(A2). Other methods include clustering such as t-SNE (t-Distributed Stochastic Neighbor Embedding) shown in fig. \ref{fig:XAIMath}(B) and subspace-related methods, for example correlation-based Singular Vector Canonical Correlation Analysis (SVCCA) \cite{SVCCARaghu} is used to find the significant directions in the subspace of input for accurate prediction, as shown in figure \ref{fig:XAIMath}(C). Information theory has been used to study interpretability by considering Information Bottleneck principle \cite{information_bottleneck_principle,DBLP:journals/corr/Shwartz-ZivT17}. The rich ways in which mathematical structures add to the interpretability pave ways to a comprehensive view of the interpretability of algorithms, hopefully providing a ground for unifying the different views under a coherent framework in the future. Fig. \ref{fig:overviewMath} provides an overview of ideas under this category.

\textit{\thesubsection .\begin{sss}\end{sss} Pre-defined Model}

To study a system of interest, especially complex systems with not well-understood behaviour, mathematical formula such as parametric models can help simplify the tasks. With a proper hypothesis, relevant terms and parameters can be designed into the model. Interpretation of the terms come naturally if the hypothesis is either consistent with available knowledge or at least developed with good reasons. When the systems are better understood, these formula can be improved by the inclusion of more complex components. In the medical field (see later section), an example is \textit{kinetic modelling}. Machine learning can be used to compute the parameters defined in the models. Other methods exist, such as integrating commonly available methodologies with subject specific contents etc. For example, Generative Discriminative Models \cite{10.1007/978-3-030-00931-1_62}, combining ridge regression and least square method to handle variables for analyzing Alzheimer's disease and schizophrenia.

\textit{Linearity}. The simplest interpretable pre-defined model is the linear combination of variables \(y=\Sigma_i a_i x_i\) where \(a_i\) is the degree of how much \(x_i\) contributes to the prediction \(y\). A linear combination model with \(x_i\in\{0,1\}\) has been referred to as the \textit{additive feature attribution method} \cite{NIPS2017_7062}. If the model performs well, this can be considered highly interpretable. However, many models are highly non-linear. In such cases, studying interpretability via linear properties (for example, using linear probe; see below) are useful in several ways, including the ease of implementation. When linear property appears to be insufficient, non-linearity can be introduced; it is typically not difficult to replace the linear component \(\vec{w}\cdot\vec{a}\) within the system with a non-linear version \(f(\vec{w},\vec{a})\). 

A linear probe is used in \cite{alain2016understanding} to extract information from each layer in a neural network. More technically, assume we have deep learning classifier \(F(x)\in [0,1]^D\) where \(F_i(x)\in [0,1]\) is the probability that input x is classified into class \(i\) out of \(D\) classes. Given a set of features \(H_k\) at layer \(k\) of a neural network, then the linear probe \(f_k\) at layer \(k\) is defined as a linear classifier \(f_k:H_k\rightarrow [0,1]^D\) i.e. \(f(h_k)=softmax(Wh_k+b)\). In another words, the probe tells us how well the information from only layer \(k\) can predict the output, and each of this predictive probe is a linear classifier by design. The paper then shows plots of the error rate of the prediction made by each \(f_k\) against \(k\) and demonstrates that these linear classifiers generally perform better at deeper layer, that is, at larger \(k\). 

\textit{General Additive Models}. Linear model is generalized by the Generalized Additive Model (GAM) \cite{hastie1986,10.1145/2339530.2339556} with standard form \(g(E[y])=\beta_0+\Sigma f_j(x_j)\) where \(g\) is the \textit{link function}. The equation is general, and specific implementations of \(f_j\) and link function depend on the task. The familiar General Linear Model (GLM) is GAM with the specific implementation of linear \(f_j\) and \(g\) is the identity. Modifications can be duly implemented. As a natural extension to the model, interaction terms between variables \(f_{ij}(x_{i},x_{j})\) are used \cite{10.1145/2487575.2487579}; we can certainly extend this indefinitely. ProtoAttend \cite{DBLP:journals/corr/abs-1902-06292} uses probabilities as weights in the linear component of the NN. Such model is considered inherently interpretable by the authors. In the medical field, see \cite{10.1007/978-3-030-00931-1_62, DBLP:journals/corr/abs-1806-07237,HAUFE201496,10.1145/2783258.2788613}.

\textit{Content-subject-specific model}. Some algorithms are considered obviously interpretable within its field. Models are designed based on existing knowledge or empirical evidence, and thus interpretation of the models is innately embedded into the system. ML algorithms can then be incorporated in rich and diverse ways, for example, through parameter fitting. The following lists just a few works to illustrate the usage diversity of ML algorithms. Deep Tensor Neural Network is used for quantum many-body systems \cite{QuantumChem}. Atomistic neural network architecture for quantum chemistry is used in \cite{QuantumChem2019}, where each atom is like a node in a graph with a set of feature vectors. The specifics depend on the neural network used, but this model is considered inherently interpretable. Neural network has been used for programmable wireless environments (PWE) \cite{DBLP:journals/corr/abs-1905-02495}. TS approximation \cite{10.1007/978-3-030-21920-8_15} is a fuzzy network approximation of other neural networks. The approximate fuzzy system is constructed with choices of components that can be adapted to the context of interpretation. The paper itself uses sigmoid-based membership function, which it considers interpretable. A so-called model-based reinforcement learning is suggested to be interpretable after the addition of high level knowledge about the system that is realized as Bayesian structure \cite{DBLP:journals/corr/abs-1907-04902}. 

\textit{Challenges and Future Prospects}. The challenge of formulating the ``correct'' model exists regardless of machine learning trend. It might be interesting if a system is found that is fundamentally operating on a specific machine learning model. Backpropagation-based deep NN (DNN) itself is inspired by the brain, but they are not operating at fundamental level of similarity (nor is there any guarantee that such model exists). When interpretability is concerned, having fundamental similarity to real, existing systems may push forward our understanding of machine learning model in unprecedented ways. Otherwise, in the standard uses of machine learning algorithm, different optimization paradigms are still being discovered. Having optimization paradigm that is specialized for specific models may be contribute to a new aspect of interpretable machine learning.

\textit{\thesubsection .\begin{sss}\end{sss} Feature Extraction}
%\subsubsection{Feature Extraction} \label{featureEx}

We give an intuitive explanation via a hypothetical example of a classifier for heart-attack prediction. Given, say, 100-dimensional features including eating pattern, job and residential area of a subject. A kernel function can be used to find out that the strong predictor for heart attack is a 100-dimensional vector which is significant in the following axes: eating pattern, exercise frequency and sleeping pattern. Then, this model is considered interpretable because we can link heart-attack risk with healthy habits rather than, say socio-geographical factors. More information can be drawn from the next most significant predictor and so on. 

\textit{Correlation}. The methods discussed in this section include the use of correlation in a general sense. This will naturally include covariance matrix and correlation coefficients after transformation by kernel functions. A kernel function transforms high-dimensional vectors such that the transformed vectors better distinguish different features in the data. For example, the Principal Component Analysis transforms vectors into the principal components (PC) that can be ordered by the eigenvalues of singular-value-decomposed (SVD) covariance matrix. The PC with the highest eigenvalue is roughly the most informative feature. Many kernel functions have been introduced, including the Canonical Correlation Analysis (CCA) \cite{CCAOverview}. CCA provides the set of features that transforms the original variables to the pairs of canonical variables, where each pair is a pair of variables that are ``best correlated'' but not correlated to other pairs. Quoted from \cite{HAZARIKA201926}, ``such features can inherently characterize the object and thus it can better explore the insights and finer details of the problems at hand''. In the previous sections, interpretability research using correlation includes \cite{kindermans2017learning}.

SVCCA combines CCA and SVD to analyze interpretability \cite{SVCCARaghu}. Given an input dataset \(X=\{x_1,...,x_m\}\) where each input \(x_i\) is possibly multi-dimensional. Denote the activation of neuron \(i\) at layer \(l\) as \(z_i^l=(z_i^l(x_1),...,z_i^l(x_m))\). Note that one such output is defined for the entire input dataset. SVCCA finds out the relation between 2 layers of a network \(l_k=\{z_i^{l_k}|i=1,...,m_k\}\) for \(k=1,2\) by taking \(l_1\) and \(l_2\) as the input (generally, \(l_k\) does not have to be the entire layer). SVCCA uses SVD to extract the most informative components \(l'_k\) and uses CCA to transform \(l'_1\) and \(l'_2\) such that \(\bar{l}'_1=W_Xl'_1\) and \(\bar{l}'_2=W_Xl'_2\) have the maximum correlation \(\rho=\{\rho_1,...,\rho_{min(m_1,m_2)}\}\). One of the SVCCA experiments on CIFAR-10 demonstrates that only 25 most-significant axes in \(l'_k\) are needed to obtain nearly the full accuracy of a full-network with 512 dimensions. Besides, the similarity between 2 compared layers is defined to be \(\bar{\rho}=\frac{1}{min(m_1,m_2)}\Sigma_i\rho_i\).

The successful development of generative adversarial networks (GAN) \cite{NIPS2014_5423,10.5555/3305381.3305404, Zhu_2017_ICCV} for generative tasks have spawned many derivative works. GAN-based models have been able to generate new images not distinguishable from synthetic images and perform many other tasks, including transferring style from one set of images to another or even producing new designs for products and arts. Studies related to interpretabilities exist. For example \cite{8803129} uses encoder-decoder system to perform multi-stage PCA. Generative model is used to show that natural image distribution modelled using probability density is fundamentally difficult to interpret \cite{DBLP:journals/corr/abs-1901-01499}. This is demonstrated through the use of GAN for the estimation of image distribution density. The resulting density shows preferential accumulation of density of images with certain features (for examples, images featuring small object with few foreground distractions) in the pixel space. The paper then suggests that interpretability is improved once it is embedded in the deep feature space, for example, from GAN. In this sense, the interpretability is offered by better correlation between the density of images with the correct identification of the objects. Consider also the GAN-based works they cite.

\textit{Clustering}. Algorithm such as t-SNE has been used to cluster input images based on their activation of neurons in a network \cite{DBLP:journals/corr/NguyenYC16, tSNECNN}. The core idea relies on the distance between objects being considered. If the distance between two objects are short in some measurement space, then they are similar. This possibly appeals to the notion of human learning by the \textit{Law of Association}. It differs from correlation-based method which provides some metrics that relate the change of one variable with another, where the two related objects can originate from completely different domains; clustering simply presents their similarity, more sensibly in similar domain or in the subsets thereof. In \cite{tSNECNN}, the activations \(\{f_{fc7}(x)\}\) of 4096-dimensional layer fc7 in the CNN are collected over all input \(\{x\}\). Then \(\{f_{fc7}(x)\}\) is fed into t-SNE to be arranged and embedded into two-dimension for visualization (each point then is visually represented by the input image \(x\)). Activation atlases are introduced in \cite{olahactivatlas}, which similarly uses t-SNE to arrange some activations \(\{f_{act}(x)\}\), except that each point is represented by the average activations of feature visualization. In meta-material design \cite{metamaterial}, design pattern and optical responses are encoded into latent variables to be characterized by Variational Auto Encoder (VAE). Then, t-SNE is used to visualize the latent space. 

In the medical field (also see later section), we have \cite{GroupINN}, \cite{DBLP:journals/corr/abs-1807-06843} (uses Laplacian Eigenmap for interpretability) \cite{10.1007/978-3-030-00928-1_73} (introduces a low-rank representation method for Autistic Spectrum Diagnosis).

\textit{Challenges and Future Prospects}. This section exemplifies the difficulty in integrating mathematics and human intuition. Having extracted "relevant" or "significant" features, sometimes we are left with still a combination of high dimensional vectors. Further analysis comes in the form of correlations or other metrics that attempt to show similarities or proximity. The interpretation may stay as mathematical artifact, but there is a potential that separation of concepts attained by these methods can be used to reorganize a black-box model from within. It might be an interesting research direction that lacks justification in terms of real-life application: however, progress in unraveling black-boxes may be a high-risk high-return investment.

%\subsubsection{Sensitivity} \label{section2:sensitivity}
\textit{\thesubsection .\begin{sss}\end{sss} Sensitivity}

We group together methods that rely on localization, gradients and perturbations under the category of “sensitivity”. These methods rely on the notion of small changes \(dx\) in calculus and the neighborhood of a point in metric spaces.

\textit{Sensitivity to input noises or neighborhood of data points}. Some methods rely on the locality of some input \(x\). Let a model \(f(.)\) predicts \(f(x)\) accurately for some \(x\). Denote \(x+\delta\) as a slightly noisy version of \(x\). The model is locally faithful if \(f(x+\delta)\) produces correct prediction, otherwise, the model is unfaithful and clearly such instability reduces its reliability. Reference  \cite{DBLP:journals/corr/FongV17} introduces \textit{meta-predictors} as interpretability methods and emphasizes the importance of the variation of input \(x\) to neural network in explaining a network. They define \textit{explanation} and \textit{local explanation} in terms of the response of blackbox \(f\) to some input. Amongst many of the studies conducted, they provide experimental results on the effect of varying input such as via deletion of some regions in the input. Likewise, when random pixels of an image are deleted (hence the data point is shifted to its neighborhood in the feature space) and the resulting change in the output is tested \cite{DBLP:journals/corr/ShrikumarGK17}, pixels that are important to the prediction can be determined. In text classification, \cite{alvarez-melis-jaakkola-2017-causal} provides “explanations”  in the form of partitioned graphs. The explanation is produced in three main steps, where the first step involves sampling perturbed versions of the data using VAE. 

Testing with Concept Activation Vectors (TCAV) has also been introduced as a technique to interpret the low-level representation of neural network layer \cite{conf_icml_KimWGCWVS18}. First, the concept activation vector (CAV) is defined. Given input \(x\in\mathbb{R}^n\) and a feedforward layer \(l\) having \(m\) neurons, the activation at that layer is given by \(f_l:\mathbb{R}^n\rightarrow\mathbb{R}^m\). If we are interested in the concept \(C\), for example “striped” pattern, then, using TCAV, we supply a set \(P_C\) of examples corresponding to “striped” pattern (zebra, clothing pattern etc) and the negative examples \(N\). This collection is used to train a binary classifier \(v_C^l\in\mathbb{R}^m\) for layer \(l\) that partitions \(\{f_l(x):x\in P_C\}\) and \(\{f_l(x):x\in N\}\). In another words, a kernel function extracts features by mapping out a set of activations that has relevant information about the “stripe”-ness. CAV is thus defined as the normal vector to the hyperplane that separates the positive examples from the negative ones, as shown in fig. \ref{fig:XAIMath}(A1). It then computes directional derivative \(S_{v,k,l}(x)=\nabla h_{l,k}(f_l(x))\cdot v_C^l\) to obtain the sensitivity of the model w.r.t. the concept \(C\), where \(h_{l,k}\) is the logit function for class \(k\) of \(C\) for layer \(l\). 

LIME \cite{10.1145/2939672.2939778} optimizes over models \(g\in G\) where \(G\) is a set of interpretable models \(G\) by minimizing locality-aware loss and complexity. In another words, it seeks to obtain the optimal model \(\xi(x)=argmin_{g\in G}\ L(f,g,\pi_x)+\Omega (g)\) where \(\Omega\) is the complexity and \(f\) is the true function we want to model. An example of the loss function is \(L(f,g,\pi_x)=\Sigma_{z,z'\in Z} \pi_x (z)[f(x)-g(z')]^2\) with \(\pi_x (z)\) being, for example, Euclidean distance and \(Z\) is the vicinity of \(x\). From the equation, it can be seen that the desired \(g\) will be close to \(f\) in the vicinity \(Z\) of \(x\), because \(f(z)\approx g(z')\) for \(z,z'\in Z\). In another words, noisy inputs \(z,z'\) do not add too much losses. 

Gradient-based explanation vector \(\xi(x_0)=\frac{\partial}{\partial x}P(Y\neq g(x_0) | X = x)\) is introduced by \cite{10.5555/1756006.1859912} for Bayesian classifier \(g(x)=argmin_{c\in\{i,...,C\}}\ P(Y\neq c|X=x)\), where \(x,\xi\) are d-dimensional. For any \(i=1,...,d\), high absolute value of \([\xi(x_0)]_i\) means that component \(i\) contributes significantly to the decision of the classifier. If it is positive, the higher the value is, the less likely \(x_0\) contributes to decision \(g(x_0)\). 

ACE algorithm \cite{ACE_NIPS2019_9126} uses TCAV to compute saliency score and generate super-pixels as explanations. Grad-CAM \cite{DBLP:journals/corr/SelvarajuDVCPB16} is a saliency method that uses gradient for its sensitivity measure. In \cite{10.5555/3305381.3305576}, \textit{influence function} is used. While theoretical, the paper also practically demonstrates how understanding the underlying mathematics will help develop perturbative training point for adversarial attack.

\textit{Sensitivity to dataset}. A model is possibly sensitive to the training dataset \(\{x_i\}\) as well. Influence function is also used to understand the effect of removing \(x_i\) for some \(i\) and shows the consequent possibility of adversarial attack \cite{10.5555/3305381.3305576}. Studies on adversarial training examples can be found in the paper and its citations, where seemingly random, insignificant noises can degrade machine decision considerably. The \textit{representer theorem} is introduced for studying the extent of effect \(x_i\) has on a decision made by a deep NN \cite{10.5555/3327546.3327602}. 

\textit{Challenges and Future Prospects}. There seems to be a concern with locality and globality of the concepts. As mentioned in \cite{conf_icml_KimWGCWVS18}, to achieve \textit{global quantification} for interpretability, explanation must be given for a set of examples or the entire class rather than ``just explain individual data inputs". As a specific example, there may be a concern with the globality of TCAV. From our understanding, TCAV is a perturbation method by the virtue of stable continuity in the usual derivative and it is global because the whole subset of dataset with label \(k\) of concept \(C\) has been shown to be well-distinguished by TCAV. However, we may want to point out that despite their claim to globality, it is possible to view the success of TCAV as local, since it is only ``global” within each label \(k\) rather than within all dataset considered at once.

From the point of view of image processing, the neighborhood of a data point (an image) in the feature space poses a rather subtle question; also refer to fig. {\ref{fig:overviewMath}(C)} for related illustration. For example, after rotating and stretching the image or deleting some pixels, how does the position of the image in the feature space change? Is there any way to control the effect of random noises and improve robustness of machine prediction in a way that is sensible to human's perception? The transition in the feature space from one point to another point that belongs to different classes is also unexplored.

On a related note, gradients have played important roles in formulating interpretability methods, be it in image processing or other fields. Current trend recognizes that regions in the input space with significant gradients provide interpretability. Deforming these regions quickly degrades the prediction; conversely, the particular values at these regions are important to the reach a certain prediction. This is helpful, since calculus exists to help analyse gradients. However, this has shown to be disruptive as well. For example, imperceptible noises can degrade prediction drastically (see \textit{manipulation of explanations} under the section \textit{Risk of Machine Interpretation in Medical Field}). Since gradient is also in the core of loss optimization, it is a natural target for further studies.

\textit{\thesubsection .\begin{sss}\end{sss} Optimization}
%\subsubsection{Optimization}

We have described several researches that seek to attain interpretability via optimization methods. Some have optimization at the core of their algorithm, but the interpretability is left to visual observation, while others optimize interpretability mathematically. 

\textit{Quantitatively maximizing interpretability}. To approximate a function  \(f\), as previously mentioned, LIME \cite{10.1145/2939672.2939778} performs optimization by finding optimal model \(\xi\in G\) so that \(f(z)\approx \xi(z')\) for \(z,z'\in Z\) where \(Z\) is the vicinity of \(x\), so that local fidelity is said to be achieved. Concurrently the complexity \(\Omega(\xi)\) is minimized. Minimized \(\Omega\) means the model’s interpretability is maximized. MUSE \cite{museHima} takes in blackbox model, prediction and user-input features to output decision sets based on optimization w.r.t fidelity, interpretability and unambiguity. The available measures of interpretability that can be optimized include \textit{size},\ \textit{featureoverlap} etc (refer to table 2 of its appendix).

\textit{Activation Optimization}.  Activation optimizations are used in research works such as \cite{Olah_2017,visualization_techreport,DBLP:journals/corr/NguyenYC16,DBLP:journals/corr/YosinskiCNFL15} as explained in a previous section. The interpretability relies on direct observation of the neuron-activation-optimized images. While the quality of the optimized images are not evaluated, the fact that parts of coherent images emerge with respect to a (collection of) neuron(s) does demonstrate some organization of information in the neural networks.  

\subsection{Other Perspectives to Interpretability}
There are many other concepts that can be related to interpretability. Reference \cite{DBLP:journals/corr/SelvarajuDVCPB16} conducted experiments to test the improvements of human performance on a task after being given explanations (in the form of visualization) produced by ML algorithms. We believe this might be an exemplary form of interpretability evaluation. For example, we want to compare machine learning algorithms \(ML_A\) with \(ML_B\). Say, human subjects are given difficult classification tasks and attain a baseline \(40\%\) accuracy. Repeat the task with different set of human subjects, but they are given explanations churned out by \(ML_A\) and \(ML_B\). If the accuracies attained are now \(50\%\) and \(80\%\) respectively, then \(ML_B\) is more interpretable.

Even then, if human subjects cannot really explain why they can perform better with the given explanations, then the interpretability may be questionable. This brings us to the question of what kind of interpretability is necessary in different tasks and certainly points to the possibility that there is no need for a unified version of interpretability. 

\begin{table*}[ht]
\caption{(continued from table \ref{tab:tablepart1}) List of journal papers arranged according to the interpretability methods used, how interpretability is presented or the suggested means of interpretability.}
\begin{center}
\begin{tabular}{lcc|c|c|c}
    \hline
	Methods & \multicolumn{2}{c}{HSI\ \ \ \ ANN} & \multicolumn{2}{c}{Mechanism} & \\
    \hline
    Linear probe \cite{alain2016understanding} & \xmark & \checkmark & \multicolumn{2}{c|}{} & \multirow{42}{*}{\rotatebox[origin=c]{-90}{Interpretability via Mathematical Structure}} \\
    Regression based on CNN \cite{DBLP:journals/corr/abs-1806-07237} & \xmark & \checkmark & \multicolumn{2}{c|}{} & \\
    Backwards model for interpretability of linear models \cite{HAUFE201496} & \xmark & \xmark & \multicolumn{2}{c|}{} & \\
    GDM (Generative Discriminative Models): ridge regression + least square \cite{10.1007/978-3-030-00931-1_62} & \xmark & \xmark & \multicolumn{2}{c|}{} & \\
	GAM, GA\(^2\)M (Generative Additive Model) \cite{10.1145/2783258.2788613,hastie1986,10.1145/2339530.2339556} & \xmark & \xmark & \multicolumn{2}{c|}{} & \\
	ProtoAttend \cite{DBLP:journals/corr/abs-1902-06292} & \xmark & \checkmark & \multicolumn{2}{c|}{} & \\
	Other content-subject-specific models: & N.A. & N.A. & \multicolumn{2}{c|}{Pre-defined models} & \\
	+ Kinetic model for CBF (cerebral blood flow) \cite{10.1007/978-3-030-00928-1_4} & N.A. & \checkmark & \multicolumn{2}{c|}{} & \\
	+ CNN for PK (Pharmacokinetic) modelling \cite{pmid30182792} & N.A. & \checkmark & \multicolumn{2}{c|}{} & \\
	+ CNN for brain midline shift detection \cite{10.1007/978-3-030-33850-3_4} & N.A. & \checkmark & \multicolumn{2}{c|}{} & \\
	+ Group-driven RL (reinforcement learning) on personalized healthcare \cite{10.1007/978-3-030-00928-1_67} & N.A. & \checkmark & \multicolumn{2}{c|}{} & \\
	+ Also see \cite{QuantumChem,QuantumChem2019,DBLP:journals/corr/abs-1905-02495,10.1007/978-3-030-21920-8_15,DBLP:journals/corr/abs-1907-04902} & N.A. & \checkmark & \multicolumn{2}{c|}{} & \\

	\hhline{-|-|-|-|-|~|}
    PCA (Principal Components Analysis), SVD (Singular Value Decomposition) & N.A. & N.A. & \multirow{9}{*}{Correlation} & \multirow{18}{*}{\rotatebox[origin=c]{-90}{Feature Extraction}} & \\
    CCA (Canonical Correlation Analysis) \cite{CCAOverview} & \xmark & \xmark &  &  & \\
    SVCCA (Singular Vector Canonical Correlation Analysis) \cite{SVCCARaghu} = CCA+SVD  & \xmark & \checkmark &  &  & \\    
    F-SVD (Frame Singular Value Decomposition) \cite{HAZARIKA201926} on electromyography data & \xmark & \xmark &  &  & \\    
    DWT (Discrete Wavelet Transform) + Neural Network \cite{KOCADAGLI2017419} & \xmark & \checkmark &  &  & \\    
    MODWPT (Maximal Overlap Discrete Wavelet Package Transform) \cite{ZHANG2019240} & \xmark & \xmark &  &  & \\    
    GAN-based Multi-stage PCA \cite{8803129} & \checkmark & \xmark &   &  & \\
    Estimating probability density with deep feature embedding \cite{DBLP:journals/corr/abs-1901-01499} & \xmark & \checkmark &   &  & \\   
	\hhline{%
	!{\leaders\hbox{\hdashrule{0.1 cm}{0.2pt}{1pt}}\hfil}%
	!{\leaders\hbox{\hdashrule{0.1 cm}{0.2pt}{1pt}}\hfil}%
	!{\leaders\hbox{\hdashrule{0.1 cm}{0.2pt}{1pt}}\hfil}%
	!{\leaders\hbox{\hdashrule{0.1 cm}{0.2pt}{1pt}}\hfil}%
	!{}%
	}	
	
   t-SNE (t-Distributed Stochastic Neighbour Embedding) \cite{DBLP:journals/corr/NguyenYC16}  & \xmark & \checkmark & \multirow{10}{*}{Clustering} & & \\  
   + t-SNE on CNN \cite{tSNECNN} & \xmark & \checkmark &  & & \\  
   + t-SNE, activation atlas on GoogleNet \cite{olahactivatlas} & \xmark & \checkmark &  & & \\  
   + t-SNE on latent space in meta-material design \cite{metamaterial} & \xmark & \checkmark &  & & \\  
   + t-SNE on genetic data \cite{doi:10.1142/S0219720017500172} & \xmark & \checkmark &  & & \\ 
   + mm-t-SNE on phenotype grouping \cite{pmid25350393} & \xmark & \checkmark &  & & \\ 
	Laplacian Eigenmaps visualization for Deep Generative Model \cite{DBLP:journals/corr/abs-1807-06843} & \xmark & \checkmark &  & & \\
	KNN (k-nearest neighbour) on multi-center low-rank rep. learning (MCLRR) \cite{10.1007/978-3-030-00928-1_73} & \xmark & \checkmark &  & & \\
	KNN with triplet loss and \textit{query-result activation map pair} \cite{10.1007/978-3-030-02628-8_11}	& \xmark & \checkmark &  & & \\
	Group-based Interpretable NN with RW-based Graph Convolutional Layer \cite{GroupINN} & \xmark & \checkmark &  & & \\
	    
    \hhline{-|-|-|-|-|~|}
    TCAV (Testing with Concept Activation Vectors) \cite{conf_icml_KimWGCWVS18} & \checkmark & \checkmark & \multicolumn{2}{c|}{} & \\
    + RCV (Regression Concept Vectors) uses TCAV with Br score \cite{10.1007/978-3-030-02628-8_14} & \xmark & \checkmark &  \multicolumn{2}{c|}{}  & \\
    + Concept Vectors with UBS \cite{10.1007/978-3-030-33850-3_2} & \xmark & \checkmark &  \multicolumn{2}{c|}{}  & \\
    +  ACE (Automatic Concept-based Explanations) \cite{ACE_NIPS2019_9126} uses TCAV & \checkmark & \checkmark &  \multicolumn{2}{c|}{}  & \\
   
    Influence function \cite{10.5555/3305381.3305576} helps understand adversarial training points & \xmark & \checkmark &  \multicolumn{2}{c|}{}  & \\
    Representer theorem \cite{10.5555/3327546.3327602} & \xmark & \checkmark &  \multicolumn{2}{c|}{Sensitivity}  & \\
    SocRat (Structured-output Causual Rationalizer) \cite{alvarez-melis-jaakkola-2017-causal} & \xmark & \checkmark &  \multicolumn{2}{c|}{}  & \\
    Meta-predictors \cite{DBLP:journals/corr/FongV17} & \xmark & \checkmark &  \multicolumn{2}{c|}{}  & \\
    Explanation vector \cite{10.5555/1756006.1859912} & \xmark & \xmark &  \multicolumn{2}{c|}{}  & \\
    \# Also listed elsewhere: \cite{10.1145/2939672.2939778,museHima,DBLP:journals/corr/SelvarajuDVCPB16,10.5555/3305890.3306024} & N.A. & N.A. &  \multicolumn{2}{c|}{}  & \\ 
	
	\hhline{-|-|-|-|-|~|}
	
	\# Also listed elsewhere: \cite{10.1145/2939672.2939778,museHima,kindermans2017learning} etc & N.A. & N.A. &  \multicolumn{2}{c|}{Optimization}  & \\
	
	\hhline{-|-|-|-|-|~|}    
	
    CNN with separable model \cite{DBLP:journals/corr/abs-1901-08125} & \xmark & \checkmark &  \multicolumn{2}{c|}{Others} & \\
    Information theoretic: Information Bottleneck \cite{information_bottleneck_principle,DBLP:journals/corr/Shwartz-ZivT17}  & \xmark & \checkmark &  \multicolumn{2}{c|}{}  & \\
    
    \hline

    Database of methods v.s. interpretability \cite{doshivelez2017towards} & N.A. & N.A. &  \multicolumn{2}{c|}{Data Driven} & \multirow{8}{*}{\rotatebox[origin=c]{-90}{Other Persp.}}\\
    Case-Based Reasoning \cite{LAMY201942} & \checkmark & \xmark &  \multicolumn{2}{c|}{}  & \\ 
    
    \hhline{-|-|-|-|-|~|} 

    Integrated Gradients \cite{10.5555/3305890.3306024,10.1117/12.2549298} & \xmark & \checkmark &  \multicolumn{2}{c|}{Invariance} & \\
    Input invariance \cite{Kindermans2019Unreliability} & \xmark & \checkmark &  \multicolumn{2}{c|}{}  & \\ 

   \hhline{-|-|-|-|-|~|} 

    Application-based \cite{10.1007/978-3-319-55524-9_22,10.1007/978-3-319-55524-9_21}  & & &  \multicolumn{2}{c|}{} & \\
    Human-based \cite{10.5555/2891460.2891655,10.1145/2675133.2675214} & N.A. & N.A. &  \multicolumn{2}{c|}{Utilities}  & \\ 
	Function-based \cite{DBLP:journals/corr/RonnebergerFB15,DBLP:journals/corr/CicekALBR16,CAM7780688, DBLP:journals/corr/SelvarajuDVCPB16, 10.1007/978-3-030-00928-1_55, SVCCARaghu,10.1007/978-3-319-55524-9_22,10.1007/978-3-319-55524-9_21,conf_icml_KimWGCWVS18} & & &  \multicolumn{2}{c|}{}  & \\

    \hline
    
\end{tabular}
\label{tab:tablepart2}
\end{center}
\end{table*}

\textit{\thesubsection .\begin{sss}\end{sss} Data-driven Interpretability}
%\subsubsection{Data-driven Interpretability}

\textit{Data in catalogue}. A large amount of data has been crucial to the functioning of many ML algorithms, mainly as the input data. In this section, we mention works that put a different emphasize on the treatment of these data arranged in catalogue. In essence, \cite{doshivelez2017towards} suggests that we create a matrix whose rows are different real-world tasks (e.g. pneumonia detection), columns are different methods (e.g. decision tree with different depths) and the entries are the performance of the methods on some \textit{end-task}. How can we gather a large collection of entries into such a large matrix? Apart from competitions and challenges, crowd-sourcing efforts will aid the formation of such database \cite{pmid30101347,pmid27883904}. A clear problem is how multi-dimensional and gigantic such tabulation will become, not to mention that the collection of entries is very likely uncountably many. Formalizing interpretability here means we pick latent dimensions (common criteria) that human can evaluate e.g. time constraint or time-spent, cognitive chunks (defined as the basic unit of explanation, also see the definition in \cite{DBLP:journals/corr/abs-1902-00006}) etc. These dimensions are to be refined along iterative processes as more user-inputs enter the repository. 

\textit{Incompleteness}. In \cite{doshivelez2017towards}, the problem of \textit{incompleteness} of problem formulation is first posed as the issue in interpretability. Incompleteness is present in many forms, from the impracticality to produce all test-cases to the difficulty in justifying why a choice of proxy is the best for some scenarios. At the end, it suggests that interpretability criteria are to be born out of collective agreements of the majority, through a cyclical process of discoveries, justifications and rebuttals. In our opinion, a disadvantage is that there is a possibility that no unique convergence will be born, and the situation may aggravate if, say, two different conflicting factions are born, each with enough advocate. The advantage lies in the existence of strong roots for the advocacy of certain choice of interpretability. This prevents malicious intent from tweaking interpretability criteria to suit ad hoc purposes. 

\textit{\thesubsection .\begin{sss}\end{sss} Invariances}
%\subsubsection{Invariances}

\textit{Implementation invariance}. Reference \cite{10.5555/3305890.3306024} suggests implementation invariance as an axiomatic requirement to interpretability. In the paper, it is stated as the following. Define two \textit{functionally equivalent} functions as \(f_1,f_2\) so that \(f_1(x)=f_x(x)\) for any \(x\) regardless of their implementation details. Given any two such networks using attribution method, then the attribution functional \(A\) will map the importance of each component of an input to \(f_1\) the same way it does to \(f_2\). In another words, \((A[f_1](x))_j=(A[f_2](x))_j\) for any \(j=1,…,d\) where \(d\) is the dimension of the input. The statement can be easily extended to methods that do not use attribution as well.

\textit{Input invariance}. To illustrate using image classification problem, translating an image will also translate super-pixels demarcating the area that provides an explanation to the choice of classification correspondingly. Clearly, this property is desirable and has been proposed as an axiomatic invariance of a reliable saliency method. There has also been a study on the input invariance of some saliency methods with respect to translation of input \(x\rightarrow x+c\) for some \(c\) \cite{Kindermans2019Unreliability}. Of the methods studied, gradients/sensitivity-based methods \cite{10.5555/1756006.1859912} and signal methods \cite{DBLP:journals/corr/ZeilerF13,springenberg2014striving} are input invariant while some attribution methods, such as integrated gradient \cite{10.5555/3305890.3306024}, are not.

\textit{\thesubsection .\begin{sss}\end{sss} Interpretabilities by Utilities}
%\subsubsection{Interpretabilities by Utilities}

The following utilities-based categorization of interpretability is proposed by \cite{doshivelez2017towards}.

\textit{Application-based}. First, an evaluation is application-grounded if human A gives explanation \(X_A\) on a specific application, so-called the end-task (e.g. a doctor performs diagnosis) to human B, and B performs the same task. Then A has given B a useful explanation if B performs better in the task. Suppose A is now a machine learning model, then the model is highly interpretable if human B performs the same task with improved performance after given \(X_A\). Some medical segmentation works will fall into this category as well, since the segmentation will constitute a visual explanation for further diagnosis/prognosis \cite{10.1007/978-3-319-55524-9_22,10.1007/978-3-319-55524-9_21} (also see other categories of the grand challenge). Such evaluation is performed, for example, by \cite{DBLP:journals/corr/SelvarajuDVCPB16}. They proposed Grad-CAM applied on guided backpropagation (proposed by \cite{springenberg2014striving}) of AlexNet CNN and VGG. The produced visualizations are used to help human subjects in Amazon Mechanical Turks identify objects with higher accuracy in predicting VOC 2007 images. The human subjects achieved \(61.23\%\) accuracy, which is \(16.79\%\) higher than visualization provided by guided backpropagation.

\textit{Human-based}. This evaluation involves real humans and simplified tasks. It can be used when, for some reasons or another, having human A give a good explanation \(X_A\) is challenging, possibly because the performance on the task cannot be evaluated easily or the explanation itself requires specialized knowledge. In this case, a simplified or partial problem may be posed and \(X_A\) is still demanded. Unlike the application-based approach, it is now necessary to look at \(X_A\) specifically for interpretability evaluation. Bigger pool of human subjects can then be hired to give a generic valuation to \(X_A\) or create a model answer \(\hat{X}_A\) to compare \(X_A\) with, and then a generic valuation is computed. 

Now, suppose A is a machine learning model, A is more interpretable compared to another ML model if it scores better in this generic valuation. In \cite{10.5555/2891460.2891655}, a ML model is given a document containing the conversation of humans making a plan. The ML model produces a "report" containing relevant predicates (words) for the task of inferring what the final plan is. The metric used for interpretability evaluation is, for example, the percentage of the predicates that appear, compared to human-made report. 
We believe the format of human-based evaluation needs not be strictly like the above. For example, hybrid human and interactive ML classifiers require human users to nominate features for training \cite{10.1145/2675133.2675214}. Two different standard MLs can be compared to the hybrid, and one can be said to be more interpretable than another if it picks up features similar to the hybrid, assuming they perform at similarly acceptable level.

\textit{Functions-based}. Third, an evaluation is functionally-grounded if there exist proxies (which can be defined a priori) for evaluation, for example sparsity \cite{doshivelez2017towards}. Some papers \cite{DBLP:journals/corr/RonnebergerFB15,DBLP:journals/corr/CicekALBR16,CAM7780688, DBLP:journals/corr/SelvarajuDVCPB16, 10.1007/978-3-030-00928-1_55, SVCCARaghu,10.1007/978-3-319-55524-9_22,10.1007/978-3-319-55524-9_21,conf_icml_KimWGCWVS18} use metrics that rely on this evaluation include many supervised learning models with clearly defined metrics such as (1) Dice coefficients (related to visual interpretability), (2) attribution values, components of canonically transformed variables (see for example CCA) or values obtained from dimensionality reduction methods (such as components of principal components from PCA and their corresponding eigenvalues), where interpretability is related to the degree an object relates to a feature, for example, classification of a dog has high values in the feature space related to four limbs, shape of snout and paws etc. Which suitable metrics to use are highly dependent on the tasks at hand.

\section{XAI IN MEDICAL FIELD}
ML has also gained traction recently in the medical field, with large volume of works on automated diagnosis, prognosis \cite{pmid29507784}. From the grand-challenge.org, we can see many different challenges in the medical field have emerged and galvanized researches that use ML and AI methods. Amongst successful deep learning models are \cite{DBLP:journals/corr/RonnebergerFB15,DBLP:journals/corr/CicekALBR16}, using U-Net for medical segmentation. However, being a deep learning neural network, U-Net is still a blackbox; it is not very interpretable. Other domain specific methods and special transformations (denoising etc) have been published as well; consider for example\cite{10.1007/978-3-030-00928-1_4} and many other works in MICCAI publications.

In the medical field the question of interpretability is far from just intellectual curiosity. More specifically, it is pointed out that interpretabilities in the medical fields include factors other fields do not consider, including risk and responsibilities \cite{DBLP:journals/corr/abs-1902-06019,pmid7235423,InterpShadow}. When medical responses are made, lives may be at stake. To leave such important decisions to machines that could not provide accountabilities would be akin to shirking the responsibilities altogether. Apart from ethical issues, this is a serious loophole that could turn catastrophic when exploited with malicious intent. 

\begin{table}[h]
\caption{Categorization by the organs affected by the diseases. Neuro* refers to any neurological, neurodevelopmental, neurodegenerative etc diseases. The rows are arranged according to the focus of the interpretability as the following: Appl.=application, Method.=methodology, Comp.=comparison}
\begin{center}
\begin{tabular}{l|l|l}
    \hline
	Appl. & brain, neuro*  & breast \cite{10.1117/12.2549298}, lung \cite{DBLP:journals/corr/abs-1901-07031,10.1145/2783258.2788613},  \\
	& \cite{10.1007/978-3-030-00928-1_4, 10.1007/978-3-030-02628-8_12, Tang2019NatureAlz, DBLP:journals/corr/abs-1810-09945} & sleep \cite{DNNSleepStageScore}, skin \cite{10.1007/978-3-030-02628-8_13} \\
	& \cite{10.1007/978-3-030-33850-3_4, ZHANG2019240, pmid30182792, 8941679}  &  others \cite{DBLP:journals/corr/abs-1806-07237}   \\
   
    \hdashline[0.5pt/2.5pt]
    Method. & brain, neuro*   & breast \cite{10.1007/978-3-030-00934-2_29, 10.1007/978-3-030-33850-3_3, 10.1007/978-3-030-02628-8_14, 10.1007/978-3-030-33850-3_2} \\
     & \cite{10.1007/978-3-030-00934-2_34,10.1007/978-3-030-00931-1_24, DBLP:journals/corr/abs-1805-08403, Letham_2015, 10.1007/978-3-030-00931-1_62} & skin \cite{10.1007/978-3-030-02628-8_11}, heart \cite{DBLP:journals/corr/abs-1807-06843} \\
    & \cite{pmid30448611, GroupINN, HAZARIKA201926, KOCADAGLI2017419} & others \cite{10.1007/978-3-030-00928-1_55, DBLP:journals/corr/abs-1901-08125, DBLP:journals/corr/abs-1805-08403, pmid25350393} \\
	
	\hdashline[0.5pt/2.5pt]
	Comp. & brain, neuro* \cite{HAUFE201496, 10.1007/978-3-030-33850-3_1} & lung \cite{10.1007/978-3-030-33850-3_5}, sleep \cite{doi:10.1063/1.5128003} \\
	 & & skin \cite{10.1007/978-3-030-33850-3_6}, other \cite{doi:10.1142/S0219720017500172} \\
	
	\hline

\end{tabular}
\end{center}
\label{tab:smalltable}
\end{table}

Many more works have thus been dedicated to exploring explainability in the medical fields \cite{DBLP:journals/corr/abs-1905-05134,doi:10.1002/widm.1312,10.1007/978-3-030-00928-1_55}. They provide summaries of previous works \cite{DBLP:journals/corr/abs-1902-06019} including subfield-specific reviews such as \cite{Kallianos2019} for chest radiograph and sentiment analysis in medicine \cite{8621359}, or at least set aside a section to promote awareness for the importance of interpretability in the medical field \cite{doi:10.1148/radiol.2019190613}. In \cite{10.3389/fmed.2020.00100}, it is stated directly that being a black-box is a ``strong limitation" for AI in dermatology, as it is not capable of performing customized assessment by certified dermatologist that can be used to explain clinical evidence. On the other hand, the exposition \cite{pmid30790315} argues that a certain degree of opaqueness is acceptable, i.e. it might be more important that we produce empirically verified accurate results than focusing too much on how to the unravel the black-box. We recommend readers to consider them first, at least for an overview of interpretability in the medical field.

We apply categorization from the previous section to the ML and AI in the medical field. Table \ref{tab:smalltable} shows categorization obtained by tagging (1) how interpretability method is incorporated: either through direct application of existing methods, methodology improvements or comparison between interpretability methods and (2) the organs targeted by the diseases e.g. brain, skin etc. As there is not yet a substantial number of significant medical researches that address interpretability, we will refrain from presenting any conclusive trend. However, from a quick overview, we see that the XAI research community might benefit from more studies comparing different existing methods, especially those with more informative conclusion on how they contribute to interpretability.

\subsection{Perceptive Interpretability}
Medical data could come in the form of traditional 2D images or more complex formats such as NIFTI or DCOM which contain 3D images with multiple modalities and even 4D images which are time-evolving 3D volumes. The difficulties in using ML for these data include the following. Medical images are sometimes far less available in quantity than common images. Obtaining these data requires consent from the patients and other administrative barriers. High dimensional data also add complexity to data processing and the large memory space requirement might prevent data to be input without modification, random sampling or down-sizing, which may compromise analysis. Other possible difficulties with data collection and management include as left/right-censoring, patients' death due to unrelated causes or other complications etc.

When medical data is available, ground-truth images may not be “correct”. Not only do these data require some specialized knowledge to understand, the lack of comprehensive understanding of biological components complicates the analysis. For example, ADC modality of MR images and the isotropic version of DWI are in some sense derivative, since both are computed from raw images collected by the scanner. Furthermore, many CT or MRI scans are presented with skull-stripping or other pre-processing. However, without a more complete knowledge of what fine details might have been accidentally removed, we cannot guarantee that an algorithm can capture the correct features. 

\textit{\thesubsection .\begin{sss}\end{sss} Saliency}

The following articles consist of direct applications of existing saliency methods. Chexpert \cite{DBLP:journals/corr/abs-1901-07031} uses GradCAM for visualization of pleural effusion in a radiograph. CAM is also used for interpretability in brain tumour grading \cite{10.1007/978-3-030-02628-8_12}. Reference \cite{Tang2019NatureAlz} uses Guided Grad-CAM and feature occlusion, providing complementary heatmaps for the classification of Alzheimer's disease pathologies. Integrated gradient method and SmoothGrad are applied for the visualization of CNN ensemble that classifies estrogen receptor status using breast MRI \cite{10.1117/12.2549298}. LRP on DeepLight \cite{DBLP:journals/corr/abs-1810-09945} was applied on fMRI data from Human Connectome Project to generate heatmap visualization. Saliency map has also been computed using primitive gradient of loss, providing interpretability to the neural network used for EEG (Electroencephalogram) sleep stage scoring \cite{DNNSleepStageScore}. There has even been a direct comparison between the feature maps within CNN and skin lesion images \cite{10.1007/978-3-030-02628-8_13}, overlaying the scaled feature maps on top of the images as a means to interpretability. Some images correspond to relevant features in the lesion, while others appear to explicitly capture artifacts that might lead to prediction bias.

The following articles are focused more on comparison between popular saliency methods, including their derivative/improved versions. Reference \cite{doi:10.1063/1.5128003} trains an artificial neural network for the classification of insomnia using physiological network (PN). The feature relevance scores are computed from several methods, including DeepLIFT \cite{DBLP:journals/corr/ShrikumarGK17}. Comparison between 4 different visualizations is performed in \cite{10.1007/978-3-030-33850-3_1}. It shows different attributions between different methods and concluded that LRP and guided backpropagation provide the most coherent attribution maps in their Alzheimer's disease study. Basic tests on GradCAM and SHAP on dermoscopy images for melanoma classification are conducted, concluding with the need for significant improvements to heatmaps before practical deployment \cite{10.1007/978-3-030-33850-3_6}.

The following includes slightly different focus on methodological improvements on top of the visualization. Respond-CAM \cite{10.1007/978-3-030-00928-1_55} is derived from \cite{CAM7780688, DBLP:journals/corr/SelvarajuDVCPB16}, and provides a saliency-map in the form of heat-map on 3D images obtained from Cellular Electron Cryo-Tomography. High intensity in the heatmap marks the region where macromolecular complexes are present.  Multi-layer class activation map (MLCAM) is introduced in \cite{10.1007/978-3-030-00934-2_34} for glioma (a type of brain tumor) localization. Multi-instance (MI) aggregation method is used with CNN to classify breast tumour tissue microarray (TMA) image's for 5 different tasks \cite{10.1007/978-3-030-00934-2_29}, for example the classification of the histologic subtype. Super-pixel maps indicate the region in each TMA image where the tumour cells are; each label corresponds to a class of tumour. These maps are proposed as the means for visual interpretability. Also, see the activation maps in \cite{10.1007/978-3-030-00931-1_24} where interpretability is studied by corrupting image and inspecting region of interest (ROI). The autofocus module from \cite{DBLP:journals/corr/abs-1805-08403} promises improvements in visual interpretability for segmentation on pelvic CT scans and segmentation of tumor in brain MRI using CNN. It uses attention mechanism (proposed by \cite{NeuralMachineTrans}) and improves it with adaptive selection of scale with which the network "sees" an object within an image. With the correct scale adopted by the network while performing a single task, human observer analysing the network can understand that a neural network is properly identifying the object, rather than mistaking the combination of the object plus the surrounding as the object itself. 

There is also a different formulation for the generation of saliency maps  \cite{10.1007/978-3-030-33850-3_3}. It defines a different softmax-like formula to extract signals from DNN for visual justification in classification of breast mass (malignant/benign). Textual justification is generated as well. 

\textit{\thesubsection .\begin{sss}\end{sss} Verbal}

In \cite{10.1145/2783258.2788613}, a rule-based system could provide the statement \textit{“has asthma \(\rightarrow\) lower risk”}, where risk here refers to death risk due to pneumonia. Likewise, \cite{Letham_2015} creates a model called \textit{Bayesian Rule Lists} that provides such statements for stroke prediction. Textual justification is also provided in the LSTM-based breast mass classifier system \cite{10.1007/978-3-030-33850-3_3}. The \textit{argumentation theory} is implemented in the machine learning training process \cite{8941679}, extracting arguments or decision rules as the explanations for the prediction of stroke based on the Asymptomatic Carotid Stenosis and Risk of Stroke (ACSRS) dataset.

One should indeed look closer at the interpretability in \cite{10.1145/2783258.2788613}. Just as many MLs are able to extract some humanly non-intuitive pattern, the rule-based system seems to have captured the strange link between asthma and pneumonia. The link becomes clear once the actual explanation based on real situation is provided: a pneumonia patient which also suffers from asthma is often sent directly to the Intensive Care Unit (ICU) rather than a standard ward. Obviously, if there is a variable ICU=0 or 1 that indicates admission to ICU, then a better model can provide more coherent explanation "\textit{asthma\(\rightarrow\)ICU\(\rightarrow\)lower risk}". In the paper, the model appears not to identify such variable. We can see that interpretability issues are not always clear-cut. 

Several researches on Visual Question Answering in the medical field have also been developed. The initiative by ImageCLEF \cite{imgclef2018,imgclef2019} appears to be at its center, though VQA itself has yet to gain more traction and successful practical demonstration in the medical sector before widespread adoption.

\textit{Challenges and Future Prospects} for perceptive interpretability in medical sector. In many cases, where saliency maps are provided, they are provided with insufficient evaluation with respect to their utilities within the medical practices. For example, when providing importance attribution to a CT scan used for lesion detection, are radiologists interested in heatmaps highlighting just the lesion? Are they more interested in looking for reasons why a haemorrhage is epidural or subdural when the lesion is not very clear to the naked eyes? There may be many such medically-related subtleties that interpretable AI researchers may need to know about.

\subsection{Interpretability via Mathematical Structure}

\textit{\thesubsection .\begin{sss}\end{sss} Pre-defined Model}
 
Models help with interpretability by providing a generic sense of what a variable does to the output variable in question, whether in medical fields or not. A parametric model is usually designed with at least an estimate of the working mechanism of the system, with simplification and based on empirically observed patterns. For example, \cite{10.1007/978-3-030-00928-1_4} uses kinetic model for the cerebral blood flow in \(ml/100g/min\) with
\begin{equation}
CBF=f(\Delta M)\frac{6000\beta\Delta M exp(\frac{PLD}{T_{1b}})}{2\alpha T_{1b}(SI_{PD})(1-exp(-\frac{\tau}{T_{1b}}))}
\end{equation}
which depends on perfusion-weighted image \(\Delta M\) obtained from the signal difference between labelled image of arterial blood water treated with RF pulses and the control image. This function is incorporated in the loss function in the training pipeline of a fully convolutional neural network. At least, an interpretation can be made partially: the neural network model is designed to denoise a perfusion-weighted image (and thus improve its quality) by considering CBF. How the network “understands” the CBF is again an interpretability problem of a neural network which has yet to be resolved. 

There is an inherent simplicity in the interpretability of models based on linearity, and thus they have been considered obviously interpretable as well; some examples include linear combination of clinical variables \cite{10.1007/978-3-030-00931-1_62}, metabolites signals for MRS \cite{DBLP:journals/corr/abs-1806-07237} etc. Linearity in different models used in the estimation of brain states is discussed in \cite{HAUFE201496}, including how it is misinterpreted. It compares what it refers to as forward and backward models and then suggested improvement on linear models. In \cite{10.1145/2783258.2788613}, a logistic regression model picked up a relation between asthma and lower risk of pneumonia death, i.e. asthma has a negative weight as a risk predictor in the regression model. Generative Discriminative Machine (GDM) combines ordinary least square regression and ridge regression to handle confounding variables in Alzheimer’s disease and schizophrenia dataset \cite{10.1007/978-3-030-00931-1_62}. GDM parameters are said to be interpretable, since they are linear combinations of the clinical variables. Deep learning has been used for PET pharmacokinetic (PK) modelling to quantify tracer target density \cite{pmid30182792}. CNN has helped PK modelling as a part of a sequence of processes to reduce PET acquisition time, and the output is interpreted with respect to the golden standard PK model, which is the linearized version of Simplified Reference Tissue Model (SRTM). Deep learning method is also used to perform parameters fitting for Magnetic Resonance Spectroscopy (MRS) \cite{DBLP:journals/corr/abs-1806-07237}. The parametric part of the MRS signal model specified, \(x(t)=\Sigma a_m x_m(t)e^{\Delta\alpha_m t+2\pi i\Delta f_m t}\), consists of linear combination of metabolite signals \(x_m(t)\). The paper shows that the error measured in SMAPE (symmetric mean absolute percentage error) is smallest for most metabolites when their CNN model is used. In cases like this, clinicians may find the model interpretable as long as the parameters are well-fit, although the neural network itself may still not be interpretable. 

The models above use linearity for studies related to brain or neuro-related diseases. Beyond linear models, other brain and neuro-systems can be modelled with relevant subject-content knowledge for better interpretability as well. Segmentation task for the detection of brain midline shift is performed using using CNN with standard structural knowledge incorporated \cite{10.1007/978-3-030-33850-3_4}. A template called \textit{model-derived age norm} is derived from mean values of sleep EEG features of healthy subjects \cite{pmid30448611}. Interpretability is given as the deviation of the features of unhealthy subject from the age norm. 

On a different note, reinforcement learning (RL) has been applied to personalized healthcare. In particular, \cite{10.1007/978-3-030-00928-1_67} introduces group-driven RL in personalized healthcare, taking into considerations different groups, each having similar agents. As usual, Q-value is optimized w.r.t policy \(\pi_\theta\), which can be qualitatively interpreted as the maximization of rewards over time over the choices of action selected by many participating agents in the system. 

\textit{Challenges and Future Prospects}. Models may be simplifying intractable system. As such, the full potential of machine learning, especially DNN with huge number of parameters, may be under-used. A possible research direction that taps onto the hype of predictive science is as the following: given a model, is it possible to augment the model with new, sophisticated components, such that parts of these components can be identified with (and thus interpreted as) new insights? Naturally, the augmented model needs to be comparable to previous models and shown with clear interpretation why the new components correspond to insights previously missed. Do note that there are critiques against the hype around the potential of AI which we will leave to the readers.

\begin{figure*}[h!]
\centering
\includegraphics[width=5.4in, trim = {0 0 0 0.5cm}]{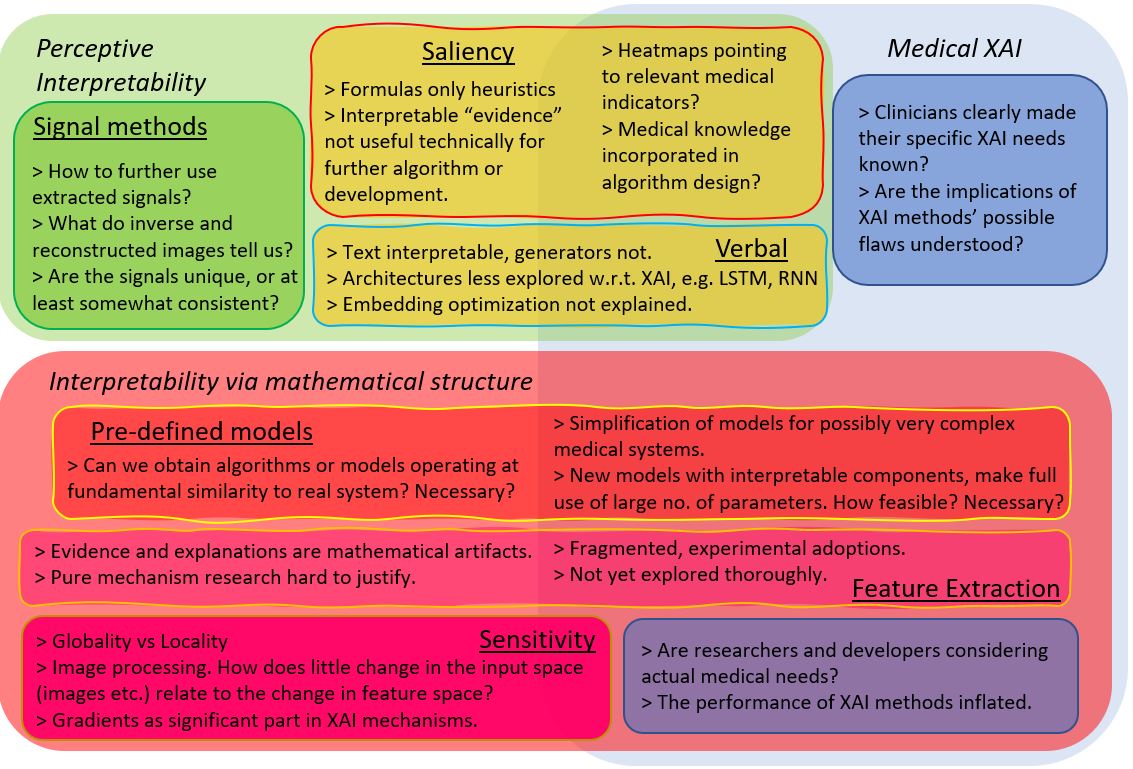}
\caption{Overview of challenges and future prospects arranged in a Venn diagram.}
\label{fig:OverviewChallenge}
\end{figure*}

\textit{\thesubsection .\begin{sss}\end{sss} Feature extraction}

Vanilla CNN is used in \cite{DBLP:journals/corr/abs-1901-08125} but it is suggested that interpretability can be attained by using a separable model. The separability is achieved by polynomial-transforming scalar variables and further processing, giving rise to weights useful for interpretation. In \cite{GroupINN}, fMRI is analyzed using correlation-based functional graphs. They are then clustered into super-graph, consisting of subnetworks that are defined to be interpretable. A convolutional layer is then used on the super-graph. For more references about neural networks designed for graph-based problems, see the paper’s citations. The following are further sub-categorization for methods that revolve around feature extraction and the evaluations or measurements (such as correlations) used to obtain the features, similar to the previous section.

\textit{Correlation}. DWT-based method (discrete wavelet transform) is used to perform feature extraction before eventually feeding the EEG data (after a series of processings) into a neural network for epilepsy classification \cite{KOCADAGLI2017419}. A \textit{fuzzy relation} analogous to correlation coefficient is then defined. Furthermore, as with other transform methods, the components (the wavelets) can be interpreted component-wise. As a simple illustration, the components for Fourier transform could be taken as how much certain frequency is contained in a time series. Reference \cite{ZHANG2019240} mentioned a host of wavelet-based feature extraction methods and introduced maximal overlap discrete wavelet package transform (MODWPT) also applied on EEG data for epilepsy classification. 

Frame singular value decomposition (F-SVD) is introduced for classifications of electromyography (EMG) data \cite{HAZARIKA201926}. It is a pipeline involving a number of processing that includes DWT, CCA and SVD, achieving around \(98\%\) accuracies on classifications between amyotrophic lateral sclerosis, myopathy and healthy subjects. Consider also CCA-based papers that are cited in the paper, in particular citations 18 to 21 for EMG and EEG signals.

\textit{Clustering}. VAE is used to obtain vectors in 64-dimensional latent dimension in order to predict whether the subjects suffer from hypertrophic cardiomyopathy (HCM) \cite{DBLP:journals/corr/abs-1807-06843}. A non-linear transformation is used to create Laplacian Eigenmap (LE) with two dimensions, which is suggested as the means for interpretability. Skin images are clustered \cite{10.1007/978-3-030-02628-8_11} for melanoma classification using k-nearest-neighbour that is customized to include CNN and triplet loss. A queried image is then compared with training images ranked according to similarity measure visually displayed as \textit{query-result activation map pair}.

t-SNE has been applied on human genetic data and shown to provide more robust dimensionality reduction compared to PCA and other methods \cite{doi:10.1142/S0219720017500172}. Multiple maps t-SNE (mm-t-SNE) is introduced by \cite{pmid25350393}, performing clustering on phenotype similarity data. 

\textit{Sensitivity}. Regression Concept Vectors (RCV) is proposed along with a metric \textit{Br} score as improvements to TCAV's concept separation \cite{10.1007/978-3-030-02628-8_14}. The method is applied on breast cancer histopathology classification problem. Furthermore, Unit Ball Surface Sampling metric (UBS) is introduced \cite{10.1007/978-3-030-33850-3_2} to address the shortcoming of \textit{Br} score. It uses neural networks for classification of nodules for mammographic images. Guidelinebased Additive eXplanation (GAX) is introduced in \cite{10.1007/978-3-030-33850-3_5} for diagnosis using CT lung images. Its pipeline includes LIME-like perturbation analysis and SHAP. Comparisons are then made with LIME, Grad-CAM and feature importance generated by SHAP.

\textit{Challenges and Future Prospects}. We observe popular uses of certain methods ingrained in specific sectors on the one hand and, on the other hand, emerging applications of sophisticated ML algorithms. As medical ML (in particular the application of recently successful DNN) is still a young field, we see fragmented and experimental uses of existing or customized interpretable methods. As medical ML research progresses, the trade-off between many practical factors of ML methods (such as ease of use, ease of interpretation of mathematical structure possibly regarded as complex) and its contribution to the subject matter will become clearer. Future research and application may benefit from a practice of consciously and consistently extracting interpretable information for further processing, and the process should be systematically documented for good dissemination. Currently, with feature selections and extractions focused on improving accuracy and performance, we may still have vast unexplored opportunities in interpretability research.

\subsection{Other Perspectives}
\textit{Data-driven}. Case-Based Reasoning (CBR) performs medical evaluation (classifications etc) by comparing a query case (new data) with similar existing data from a database. \cite{LAMY201942} combines CBR with an algorithm that presents the similarity between these cases by visually providing proxies and measures for users to interpret. By observing these proxies, the user can decide to take the decision suggested by the algorithm or not. The paper also asserts that medical experts appreciate such visual information with clear decision-support system.

\subsection{Risk of Machine Interpretation in Medical Field}
\textit{Jumping conclusion}. According to \cite{10.1145/2783258.2788613}, logical statements such as \textit{has asthma}\(\rightarrow\)\textit{lower risk} are considered interpretable. However, in the example, the statement indicates that a patient with asthma has lower risk of death from pneumonia, which might be strange without any clarification from the intermediate thought process. While human can infer that the lowered risk is due to the fact that pneumonia patients with asthma history tend to be given more aggressive treatment, we cannot always assume there is a similar humanly inferable reason behind each decision. Furthermore, interpretability method such as LRP, deconvolution and guided backpropagation introduced earlier are shown to not work for simple model, such as linear model, bringing into question their reliability \cite{kindermans2017learning}.

\section{Conclusion}
We present a survey on interpretability and explainability of ML algorithms in general, and place different interpretations suggested by different research works into distinct categories. From general interpretabilities, we apply the categorization into the medical field. Some attempts are made to formalize interpretabilities mathematically, some provide visual explanations, while others might focus on the improvement in task performance after being given explanations produced by algorithms. At each section, we also discuss related challenges and future prospects. Fig. {\ref{fig:OverviewChallenge}} provides a diagram that summarizes all the challenges and prospects.

\textit{Manipulation of explanations}. Given an image, a similar image can be generated that is perceptibly indistinguishable from the original, yet produces radically different output \cite{ghorbani2017interpretation}. Naturally, its significance attribution and interpretable information become unreliable. Furthermore, explanation can even be manipulated arbitrarily \cite{NIPS2019_9511}. For example, an explanation for the classification of a cat image (i.e. particular significant values that contribute to the prediction of cat) can be implanted into the image of a dog, and the algorithm could be fooled into classifying the dog image as a cat image. The risk in medical field is clear: even without malicious, intentional manipulation, noises can render “explanations” wrong. Manipulation of algorithm that is designed to provide explanation is also explored in \cite{Lakkaraju_2020}.

\textit{Incomplete constraints}. In \cite{10.1007/978-3-030-00928-1_4}, the loss function for the training of a fully convolutional network includes CBF as a constraint. However, many other constraints may play important roles in the mechanism of a living organ or tissue, not to mention applying kinetic model is itself a simplification. Giving an interpretation within limited constraints may place undue emphasis on the constraint itself. Other works that use predefined models might suffer similar problems \cite{10.1007/978-3-030-00931-1_62,pmid30182792,DBLP:journals/corr/abs-1806-07237}.

\textit{Noisy training data}. The so-called ground truths for medical tasks, provided by professionals, are not always absolutely correct. In fact, news regarding how AI beats human performance in medical imaging diagnosis \cite{DBLP:journals/corr/LiuGNDKBVTNCHPS17} indicates that human judgment could be brittle. This is true even of trained medical personnel. This might give rise to the classic garbage-in-garbage-out situation.

The above risks are presented in large part as a reminder of the nature of automation. It is true that algorithms have been used to extract invisible patterns with some successes. However, one ought to view scientific problems with the correct order of priority. The society should not risk over-allocating resources into building machine and deep learning models, especially since due improvements to understanding the underlying science might be the key to solving the root problem. For example, higher quality MRI scans might reveal key information not “visible” with current technology, and many models built nowadays might not be very successful because there is simply not enough detailed information contained in currently available MRI scans.

\textit{Future directions for clinicians and practitioners}. Visual and textual explanation supplied by an algorithm might seem like the obvious choice; unfortunately, the details of decision-making by algorithms such as deep neural networks are still not clearly exposed. When an otherwise reliable deep learning model provides a strangely wrong visual or textual explanation, systematic methods to probe into the wrong explanations do not seem to exist, let alone methods to correct them. A specialized education combining medical expertise, applied mathematics, data science etc might be necessary to overcome this. For now, if "interpretable" algorithms are deployed in medical practices, human supervision is still necessary. Interpretability information should be considered nothing more than complementary support for the medical practices before there is a robust way to handle interpretability.

\textit{Future directions for algorithm developers and researchers}. Before the blackbox is un-blackboxed, machine decision always carries some exploitable risks. It is also clear that a unified notion of interpretability is elusive. For medical ML interpretability, more comparative studies between the performance of methods will be useful. The interpretability output such as heatmaps should be displayed and compared clearly, including poor results. In the best case scenario, clinicians and practitioners recognize the shortcomings of interpretable methods but have a general idea on how to handle them in ways that are suitable to medical practices. In the worst case scenario, the inconsistencies between these methods can be exposed. The very troubling trend of journal publications emphasizing good results is precarious, and we should thus continue interpretability research with a mindset open to evaluation from all related parties. Clinicians and practitioners need to be given the opportunity for fair judgment of utilities of the proposed interpretability methods, not just flooded with performance metrics possibly irrelevant to the adoption of medical technology.

Also, there may be a need to shift interpretability study away from algorithm-centric studies. An authoritative body setting up the standard of requirements for the deployment of model building might stifle the progress of the research itself, though it might be the most efficient way to reach an agreement. This might be necessary to prevent damages, seeing that even corporate companies and other bodies non-academic in the traditional sense have joined the fray (consider health-tech start-ups and the implications). Acknowledging that machine and deep learning might not be fully mature for large-scale deployment, it might be wise to deploy the algorithms as a secondary support system for now and leave most decisions to the traditional methods. It might take a long time before humanity graduates from this stage, but it might be timely: we can collect more data to compare machine predictions with traditional predictions and sort out data ownership issues along the way.
% all okay $

%Subsection text here.
%
%% needed in second column of first page if using \IEEEpubid
%%\IEEEpubidadjcol
%
%\subsubsection{Subsubsection Heading Here}
%Subsubsection text here.

%\appendices
%% you can choose not to have a title for an appendix
%% if you want by leaving the argument blank
%\section{}

\section*{Acknowledgment}
This research was supported by Alibaba Group Holding Limited, DAMO Academy, Health-AI division under Alibaba-NTU Talent Program. The program is the collaboration between Alibaba and Nanyang Technological university, Singapore.

\ifCLASSOPTIONcaptionsoff
  \newpage
\fi



%\bibliographystyle{IEEEtran}
%\
%\begin{thebibliography}{1}
%\end{thebibliography}

\bibliographystyle{unsrt}
\bibliography{XAISurvey}


%\end{IEEEbiography}
%% insert where needed to balance the two columns on the last page with
%%\newpage


%\vfill


\end{document}